\documentclass{article} %
\usepackage{iclr2026_conference,times}

\usepackage{amsmath,amsfonts,bm}

\def\eqref#1{equation~\ref{#1}}

\def\1{\bm{1}}

\DeclareMathAlphabet{\mathsfit}{\encodingdefault}{\sfdefault}{m}{sl}
\SetMathAlphabet{\mathsfit}{bold}{\encodingdefault}{\sfdefault}{bx}{n}

\usepackage{hyperref}
\usepackage{url}

\usepackage{algorithm}
\usepackage{graphicx}
\usepackage{subcaption}
\usepackage{algorithmic}
\usepackage{float}
\usepackage{amsmath,amssymb,amsfonts,bm,mathtools}
\newtheorem{theorem}{Theorem}
\newtheorem{lemma}[theorem]{Lemma}
\usepackage{booktabs}
\usepackage{array}                 %
\newcolumntype{M}[1]{>{\centering\arraybackslash}m{#1}}

\usepackage{multirow}

\title{Guided Transfer Learning for Discrete Diffusion Models}

\author{ \\
\noindent \textbf{Julian Kleutgens}$^{1,2}$,
\textbf{Claudio Battiloro}$^{1}$,
\textbf{Lingkai Kong}$^{1}$,
\textbf{Benjamin Grewe}$^{2}$, \\
\textbf{Francesca Dominici}$^{1}$,
\textbf{Mauricio Tec}$^{1}$\\
$^{1}$ Harvard University, $^{2}$ ETH Zürich
}

\iclrfinalcopy %
\begin{document}

\maketitle

\begin{abstract}
Discrete diffusion models (DMs) have achieved strong performance in language and other discrete domains, offering a compelling alternative to autoregressive modeling. Yet this performance typically depends on large training datasets, challenging the performance of DMs in small-data regimes---common under real-world constraints. Aimed at this challenge, recent work in continuous DMs suggests that transfer learning via classifier ratio--based guidance can adapt a pretrained DM to a related target distribution, often outperforming alternatives such as full-weight fine-tuning on the target data. By contrast, transfer learning for discrete DMs remains  unexplored.
We address this gap by exploring practical analogues of ratio-based transfer learning for discrete DMs. Our theoretical analysis shows that a direct extension of existing ratio-based guidance is computationally prohibitive, scaling with vocabulary size. To overcome this limitation, we introduce a scheduling mechanism that yields a practical algorithm, \emph{Guided Transfer Learning} for discrete diffusion models (GTL). GTL enables sampling from a target distribution without modifying the pretrained denoiser and reduces the cost to linear scaling in vocabulary size, which in turn supports longer sequence generation.
We evaluate GTL on sequential data, including synthetic Markov chains and language modeling tasks, and provide a detailed empirical analysis of its behavior. The results highlight a clear trade-off: when target datasets are large, weight fine-tuning is often preferable, whereas GTL becomes increasingly effective as target data shrinks. Finally, we experimentally demonstrate a key failure mode of GTL: when the source and target distributions overlap poorly, the ratio-based classifier required for guidance becomes unreliable, limiting transfer performance.
\end{abstract}

\section{Introduction}

Diffusion models (DMs) \citep{ho2020denoisingdiffusionprobabilisticmodels,song2021scorebasedgenerativemodelingstochastic} have become the dominant approach for generative modeling in continuous domains like image, video, and audio  \citep{lovelace2023latentdiffusionlanguagegeneration,ho2022videodiffusionmodels, liu2023audioldmtexttoaudiogenerationlatent}.  More recently, discrete DMs for language have achieved competitive performance and, by denoising tokens in parallel and in any order, offer practical advantages for controllable infilling and editing compared to left-to-right autoregressive decoding \citep{gong2025scalingdiffusionlanguagemodels,li2022diffusionlm}.
Among existing discrete diffusion approaches, masked diffusion models (MDMs) have demonstrated remarkable performance in tasks such as reasoning~\citep{nie2025scalingmaskeddiffusionmodels,zheng2025maskeddiffusionmodelssecretly}, planning~\citep{ye2025autoregressiondiscretediffusioncomplex}, and infilling~\citep{gong2025scalingdiffusionlanguagemodels}, often performing on par with state-of-the-art auto-regressive language models (ALMs).

Training DMs require large amounts of data \citep{wang2023patchdiffusionfasterdataefficient}. And it remains an open question how to best leverage their powerful generative modeling capabilities in domains where scarse data is available \citep{zhang2025training}, a common scenario in applications like medical imaging, where privacy and limited public availability constrain dataset size \citep{xie2025meddiffftdataefficientdiffusionmodel}. This problem motivates the exploration of transfer learning techniques, where the goal is leverage a DM pretrained on a large dataset for generative modeling in a related domain where data is more scarse. One possible solution is to fine-tune the pretrained DM's denoiser on the target data \citep{Xie2023DiffFitUT}, but this becomes ineffective in very small data regimes \citep{moon2022finetuning}. To address this, recent works have proposed \textit{transfer learning} to tackle the small data regime by using classifier-based guidance (CBG)~\citep{zhong2025domainguidancesimpletransfer, ouyang2024transferlearningdiffusionmodels}, which estimates a density-ratio between the source and target domain and uses it as a diffusion guidance mechanism.  Such methods are more effective than finetuning in scenarios of extreme scarcity, as they bypass the need of updating millions of parameters \citep{moon2022finetuning, xie2023difffitunlockingtransferabilitylarge}. %
However, these methods have only been investigated in continuous domains (e.g., images). By contrast, transfer learning for discrete DMs such as language models, remains largely unexplored.

Outside of transfer learning, \citet{schiff2025simpleguidancemechanismsdiscrete, nisonoff2025unlockingguidancediscretestatespace} have explored CBG mechanisms for discrete DMs. These methods apply mainly to domains with small vocabulary size $|\mathcal{V}|$ and modest sequence length  $L$, such as biological sequences, leaving scalable CBG for language modeling as an open problem. This constaints are due to the necessity of token substitutions across all positions and vocabulary items during score computation. More precisely, naive CBG requires $\mathcal{O}(L|\mathcal{V}|)$ guidance evaluations per step, which is infeasible for long sequences and large vocabularies. In this paper, we aim to leverage CBGs for transfer learning, while overcoming the limiting small vocabulary sizes.

First, we present a theoretical analysis deriving a closed-form ratio-guided reverse transition for discrete-time diffusion and extend this principle to continuous-time, score-based discrete diffusion, covering both score-based models~\citep{lou2024discretediffusionmodelingestimating, sun2023scorebasedcontinuoustimediscretediffusion} and discrete-time diffusion models~\citep{austin2023structureddenoisingdiffusionmodels}. Such derivation is architecture- and corruption-agnostic---deriving the result from the evidence lower bound (ELBO).
Next, we address the scaling problem of CBGs for discrete DMs. For this purpose, we introduce an efficient sampler that makes guided language modeling for masked discrete DMs practical by avoiding $\mathcal{O}(L|\mathcal{V}|)$ guidance calls, concentrating ratio evaluations on planner-selected positions and a top-$n_{\text{ratio}}$ candidate set. 
\begin{figure}[t]
    \centering
    \begin{subfigure}[t]{0.59\linewidth}
        \centering
        \includegraphics[width=\linewidth]{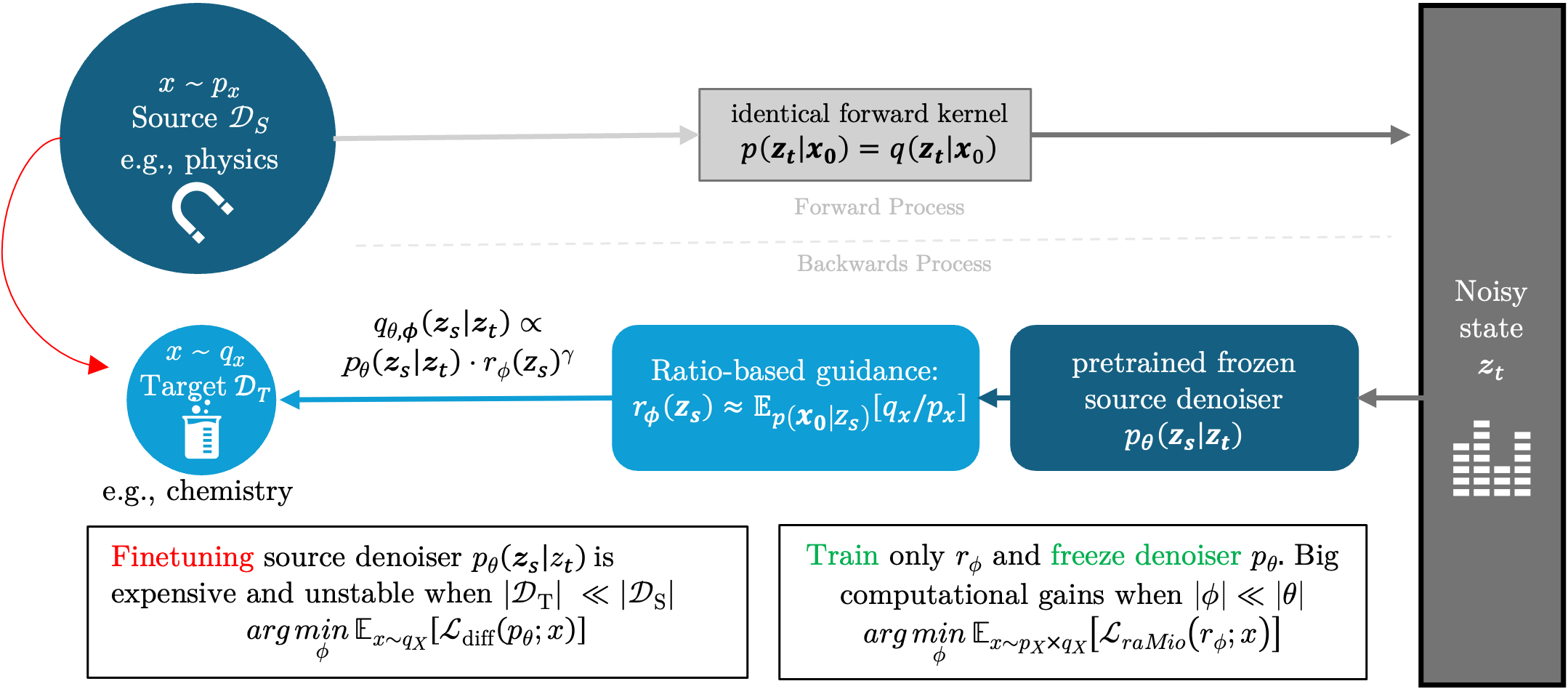}
        \label{fig:gtl_overview}
    \end{subfigure}
    \hfill
    \begin{subfigure}[t]{0.39\linewidth}
        \centering
        \includegraphics[width=\linewidth]{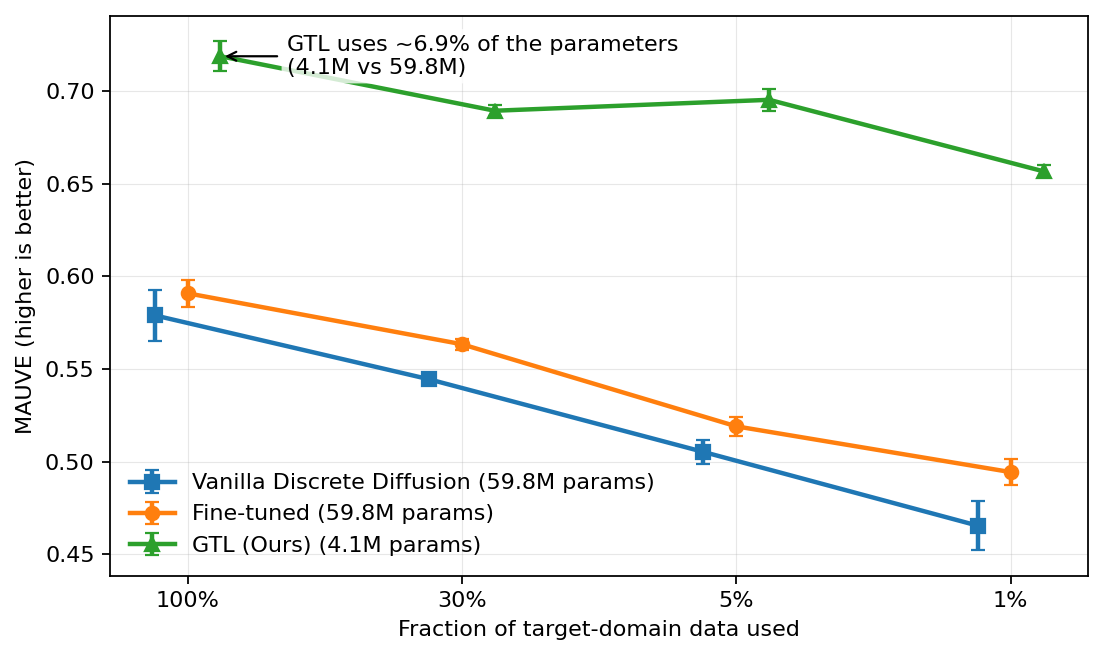}
        \label{fig:main_results}
    \end{subfigure}

    \caption{\textit{Visual summary of Guided Transfer Learning (GTL) for Discrete Diffusion and Results Preview}.
    (Left) Method overview: adapt to the target domain by reweighting the frozen source reverse transitions using a learned ratio model.
    (Right) GTL outperforms finetuning with only $\sim$7\% of the parameters. MAUVE ($\uparrow$) vs. fraction of target-domain training data (100\% = all 79,631 arXiv Physics abstracts; $1\% \approx 800$).
    GTL (green) achieves the highest MAUVE across all data regimes
    outperforming Vanilla (target-only) and Fine-tuned baselines.
    Points show mean and error bars show standard error over $3\times3$ (train$\times$sample) seeds.
    See Section~\ref{sec:exp_languagemodeling} for setup and evaluation details.}
    \label{fig:gtl_overview_and_results}
\end{figure}

Leveraging this result, we introduce a method termed Guided Transfer Learning (GTL). The core idea of GTL is to freeze a pretrained source denoiser and learn a lightweight ratio network from mixed source/target data to estimate target-to-source density ratios; at sampling time, we use this ratio with a learned planner network as guidance to reweight the denoiser’s reverse transition and generate target-domain samples without fine-tuning. This procedure allows to effectively utilize transfer learning based on CBG for discrete DMs. Previewing our results, GTL outperforms vanilla and fine-tuned diffusion across data-scarcity regimes while training only $\sim7\%$ as many parameters (Fig.~\ref{fig:gtl_overview_and_results}). In a setting common in practice—where the pretraining corpus already contains some in-domain target samples—we find that GTL performs better and yields more stable performance. Our result also reveals limitations of CBG for discrete modeling: under weak domain overlap, the interaction between guidance strength $\gamma$ and candidate pruning can amplify ratio errors. In this regime, larger $\gamma$ or small $n_{\text{ratio}}$ may destabilize denoising and hurt quality.

We summarize our contributions as follows:
\begin{enumerate}
\item We introduce GTL, an effective transfer learning framework for discrete diffusion models. GTL keeps the source denoiser fixed while training a lightweight ratio estimator on both source and target data to enable sampling from the target distribution. GTL achieves scalability by using an unmasking position for the top-$n_{\text{ratio}}$ candidates, dramatically reducing the per-step computational cost, without sacrificing performance in practice.
\item We provide a theoretical analysis and formal theorem of the proposed guided transfer learning method, demonstrating that ratio-based guidance correctly recovers the target reverse transition.
\item We validate GTL and the guided sampler on sequential data, including synthetic Markov chains and arXiv-abstract language modeling tasks.
\end{enumerate}

\section{Background}
Diffusion models are generative models defined by a forward noising process and a learned reverse denoising process. In the reverse process, a denoiser $p_{\theta}$ is trained to remove noise from latent variables $\mathbf z_t$ to recover data $x$. The forward process generates $\mathbf z_t$ by injecting noise into clean data. We consider discrete latents: $\mathbf z_t \in \mathcal V$, the set of one-hot tokens over a vocabulary $ \mathcal V$ of size $N$, $\mathcal V \coloneqq \{\mathbf z \in \{0,1\}^N : \sum_{i=1}^{N} z_i = 1\} \subset \Delta^{N-1},$ where $\Delta^{N-1}$ denotes the probability simplex over $N$ categories. Let $\text{Cat}(\cdot;\mathbf \pi)$ be the categorical distribution over $N$ classes with the probability vector $\mathbf \pi \in \Delta^{N-1}$. 
\paragraph{Discrete diffusion}
Discrete Denoising Diffusion Probabilistic Models (D3PMs)~\citep{austin2023structureddenoisingdiffusionmodels} provide a flexible framework for diffusion over discrete state spaces. The forward Markov chain is
$p(\mathbf  z_t \mid \mathbf  z_s) = \text{Cat}\!\left(\mathbf  z_t; \, Q_{t\mid s} \mathbf  z_s\right)$
where $Q_{t\mid s}\in\mathbb{R}^{N\times N}$ is a transition matrix whose $(i,j)$ entry is the probability of moving to state $i$ at time $t$ from state $j$ at time $s$.  To fit the generative model $p_{\theta}$ to the data distribution over $\mathbf x_0$, we minimize the negative ELBO (NELBO). Different choices of $Q_{t\mid s}$ induce different corruption behaviors, e.g., uniform, absorbing, or discretized-Gaussian transitions.

\paragraph{Masked (absorbing-state) diffusion.}
The absorbing-state variant or often called \emph{masked diffusion}, replaces tokens with a special \texttt{[MASK]} token~\citep{sahoo2024simpleeffectivemaskeddiffusion,shi2025simplifiedgeneralizedmaskeddiffusion}. It can be viewed as an interpolation between the clean distribution and a noise prior over the vocabulary. For a clean token $\mathbf x_0$, the forward transition at time $t$ is
\begin{align}
p(\mathbf z_t \mid \mathbf x_0) = \text{Cat}\!\left(\mathbf  z_t;\, \alpha_t \mathbf x_0 + (1-\alpha_t)\,\mathbf  m\right),
\end{align}
where $\alpha_t\!\in[0,1]$ controls the corruption schedule and $\mathbf m \in \mathcal{V}$ is one-hot vector at the special \texttt{[MASK]} token. Making use of the masking process, the backward posterior simplifies to: 
\begin{align}
&p(\mathbf z_s \mid \mathbf z_t, \mathbf x_0) = 
\begin{cases}
\operatorname{Cat}\!\left(\mathbf z_s;\, \mathbf z_t\right), & \text{if }  \mathbf z_t \neq \mathbf m,\\[6pt]
\operatorname{Cat}\!\left( \mathbf z_s;\, \dfrac{(1-\alpha_s) \mathbf m + (\alpha_s-\alpha_t) \mathbf x_0}{1-\alpha_t}\right), & \text{if } \mathbf  z_t =\mathbf  m.
\end{cases}
\end{align}
Hence, when a token is unmasked ($\mathbf x_t\neq \mathbf m$) the step is deterministic: $\mathbf x_s= \mathbf x_t$. Otherwise, the posterior is a linear interpolation between the mask distribution and the clean token.
This structure yields the simplified diffusion loss~\citep{sahoo2024simpleeffectivemaskeddiffusion,shi2025simplifiedgeneralizedmaskeddiffusion}:
\begin{align}
\mathcal{L}_{\text{diffusion}}& = \sum_{i=1}^{T}D_{\mathrm{KL}}\!\left(p(\mathbf{z}_{s(i)}\mid \mathbf{z}_{t(i)},\mathbf{x}_0)\,\big\|\, p_{\theta}(\mathbf{z}_{s(i)}\mid \mathbf{z}_{t(i)})\right)
= \sum_{i=1}^{T} \mathbb{E}_{q}\!\left[
\frac{\alpha_{t(i)} - \alpha_{s(i)}}{1 - \alpha_{t(i)}}
\log \big\langle \mathbf{x}_{\theta}(\mathbf{z}_{t(i)}), \mathbf{x}_0 \big\rangle
\right], \label{equ_bg:loss}
\end{align}
where $\mathbf{x}$ is the one-hot encoding of the clean data and $\langle \cdot,\cdot\rangle $ is the inner product. The network $\mathbf{x}_{\theta}$ predicts the clean token distribution from the noisy input $\mathbf{z}_{t(i)}$.
\paragraph{Guidance} 
Diffusion models provide strong controllability through both classifier-based and classifier-free guidance. 
These two approaches correspond to different ways of expressing the backward generative process conditioned on $y$. 
In this work, we focus on the classifier-based variant, which allows us to rearrange the conditional generative process using Bayes' rule as 
$p^{\gamma}(\mathbf{z}_s \mid \mathbf{z}_t, y) = p(y \mid \mathbf{z}_s, \mathbf{z}_t)^{\gamma} \, p(\mathbf{z}_s \mid \mathbf{z}_t) / Z$, with $Z$ being a normalizing constant. \citep{schiff2025simpleguidancemechanismsdiscrete, nisonoff2025unlockingguidancediscretestatespace}

For the extension to sequences of tokens $\mathbf{z}_s^{(1:L)}$ of length $L$, 
we assume that the distribution $p^{\gamma}\!\left(\mathbf{z}_s^{(1:L)} \mid \mathbf{z}_t^{(1:L)}, y\right)$ factorizes independently across tokens. 
This leads to the following formulation:
\begin{align}
&p^{\gamma}\!\left(\mathbf{z}_s^{(1:L)} \mid \mathbf{z}_t^{(1:L)}, y\right) = \prod_{\ell=1}^{L} 
\frac{p\!\left(y \mid \mathbf{z}_s^{(\ell)}, \mathbf{z}_t^{(1:L)}\right)^{\gamma} \, 
p\!\left(\mathbf{z}_s^{(\ell)} \mid \mathbf{z}_t^{(1:L)}\right)}
{\sum_{\tilde{\mathbf{z}}_s^{(\ell)} \in \mathcal V} 
p\!\left(y \mid \tilde{\mathbf{z}}_s^{(\ell)}, \mathbf{z}_t^{(1:L)}\right)^{\gamma} \, 
p\!\left(\tilde{\mathbf{z}}_s^{(\ell)} \mid \mathbf{z}_t^{(1:L)}\right)} \label{equ_bg:guidance}
\end{align}

Here, $\mathcal V$ denotes the vocabulary set, and  $p\!\left(y \mid \mathbf{z}_s^{(\ell)}, \mathbf{z}_t^{(1:L)}\right)$  represents the classifier probability given the sequence $\mathbf{z}_t^{(1:L)}$ with the token $\mathbf{z}_s$ placed at position $\ell$.  However, this formulation comes with the drawback of requiring $\mathcal{O}(L \cdot |\mathcal V|)$ forward passes through the classifier model, which is infeasible for long sequences and large vocabularies. Because each evaluation corresponds to a distinct single-token intervention on the full sequence, this cost is not removed by straightforward parallelization.

\paragraph{Problem Formulation}
Let $\mathcal{X}$ denote the data space, with source domain distribution $p_X$ and target domain distribution $q_X$. 
We are given source samples $\mathcal{S} = \{\mathbf{x}_i^{(1:L)}\}_{i=1}^m \sim p_X$ and target samples $\mathcal{T} = \{\mathbf{x}^{(1:L)}_i\}_{i=1}^n \sim q_X$. 
The goal of transfer learning is to train a network $p_\theta$ on $\mathcal{S}$ and adapt it to $\mathcal{T}$ to improve performance in the target domain.  
In practice, $n \ll m$, creating an imbalance between source and target data. 
A common strategy is to freeze most pretrained layers, extract features, and train new task-specific layers on $\mathcal{T}$. 
Fine-tuning relaxes this by unfreezing parameters, allowing the model to adapt more flexibly to the target distribution. In our setting, we are given a pretrained generative model $p_\theta$ trained on the source distribution $p_X$, and we aim to transfer its generative process to the target distribution $q_X$, yielding a target-domain model $q_\psi$. Note that the background section is introduced for the source domain diffusion.

\section{Guided Transfer Learning for DDMs}
We introduce \emph{Guided Transfer Learning} (GTL), a transfer framework for discrete diffusion models that adapts a pretrained source model to a target domain \emph{without} updating the denoiser. Our approach is orthogonal to fine-tuning: rather than updating the denoiser, GTL adapts sampling by reweighting the source reverse transitions with a learned density-ratio guidance model. In this section, we first present the core transfer identity for a single denoising step, then extend it to sequences and develop an efficient guided sampler.
\newline \noindent \textbf{\textit{Why not fine-tune?}} Fine-tuning adapts the source denoiser \(p_\theta\) to the target domain by optimizing the diffusion objective (e.g., \eqref{equ_bg:loss}) on target samples. This optimization becomes expensive for large language models and tends to overfit when target data are scarce, often degrading diversity. These small-data failures motivate our approach: rather than updating millions of denoiser parameters, we reuse \(p_\theta\) as-is and place the adaptation burden on a much smaller auxiliary model.  
\newline \noindent \textbf{\textit{Key Assumption.}}
This auxiliary model is meaningful only if the two domains share the same corruption mechanism.
Let \(p_X\) denote the source data distribution and \(q_X\) the target data distribution over \(\mathbf x_0\).
We assume a shared forward process,
\(
p(\mathbf z_t \mid \mathbf x_0)=q(\mathbf z_t \mid \mathbf x_0)
\)
for all \(t\), which holds whenever both domains use the same discrete forward kernel (e.g., an identical masking schedule or transition family).
Under this shared corruption, differences between the domains can be expressed at the data level through the density ratio $r(\mathbf x_0)\coloneqq q_X(\mathbf x_0)/p_X(\mathbf x_0),$ assumed well-defined on the support of interest.
This ratio quantifies how much more (or less) likely a clean sequence is under the target distribution than under the source, and it is exactly the quantity needed to reweight source reverse transitions into target reverse transitions.
\newline The next theorem makes this reweighting precise for a single reverse transition: it shows that the reverse kernel minimizing the target KL objective can be written as the source reverse kernel multiplied by a ratio-dependent weight.

\begin{theorem}\label{theorem1} Let $p$ (source) and $q$ (target) be two diffusion models that share the same forward process, \(p(\mathbf z_t\mid\mathbf x_0)=q(\mathbf z_t\mid\mathbf x_0)\) for all $t$. The reverse target domain network $q_\theta(\mathbf z_{s(i)}\mid\mathbf z_{t(i)})$ is defined by: 
\begin{align}
    \underset{\psi}{\arg\min} \; \mathbb{E}_{q}\!\left[\sum_{i=1}^{T}
    D_{\mathrm{KL}}\!\Bigl(q(\mathbf z_{s(i)}\mid\mathbf z_{t(i)},\mathbf x_0)\,\Big\|\,q_\psi(\mathbf z_{s(i)}\mid\mathbf z_{t(i)})\Bigr)\right] \label{equ_mthd:theorm1_loss}
\end{align}
Then we have
\begin{align}
    q_{\psi^\star}(\mathbf z_{s(i)}\mid\mathbf z_{t(i)}) =\frac{p(\mathbf z_{s(i)} \mid \mathbf z_{t(i)}) \mathbb{E}_{\mathbf x_0 \sim p(\cdot \mid \mathbf{z}_{s(i)})}\Bigr[\frac{q(\mathbf x_0)}{p(\mathbf x_0)} \Bigl]}{\displaystyle\sum_{\mathbf{\tilde{z}}_{s(i)}}p(\mathbf{\tilde{z}}_{s(i)}\mid \mathbf z_{t(i)})  \mathbb{E}_{\mathbf x_0 \sim p(\cdot \mid \mathbf{\tilde{z}}_{s(i)})}\Bigr[\frac{q(\mathbf x_0)}{p(\mathbf x_0)} \Bigl]} \label{equ_mthd:theorm1_solution}
\end{align}
\end{theorem}
\noindent \textbf{\textit{Practical implementation.}}
Equation~\ref{equ_mthd:theorm1_solution} (proved in Appendix~\ref{sec_apx:proof}) expresses the target reverse transition as a reweighted version of the source reverse transition. This expression suggests a direct implementation with two components: a frozen source denoiser that provides \(p(\mathbf z_{s(i)}\mid \mathbf z_{t(i)})\) and a learned ratio module that provides the ratio-dependent weight. Concretely, we approximate the source kernel with the pretrained denoiser \(p_\theta(\mathbf z_{s(i)}\mid \mathbf z_{t(i)})\), and we learn a ratio function
\( r_\phi(\mathbf z_{s(i)}) \approx  \mathbb{E}_{\mathbf x_0 \sim p(\cdot \mid \mathbf z_{s(i)})}\!\left[\frac{q(\mathbf x_0)}{p(\mathbf x_0)}\right]. \)
Together, these two networks define the GTL sampler, \(\psi^\star=[\theta,\phi]\), which generates target-domain samples while keeping the denoiser fixed. The ratio network is trained using a discrete-data loss adapted from \citet{ouyang2024transferlearningdiffusionmodels}.
Additional training details are provided in the supplementary materials.
\newline \noindent \textbf{\textit{Why ratio guidance helps in the low-data regime?}}
The key advantage of \eqref{equ_mthd:theorm1_solution} is that it eliminates the need to fine-tune the denoiser on scarce target data.
Instead, adaptation is shifted to learning a domain discriminator (or an equivalent ratio estimator) from mixed source/target samples, which can be orders of magnitude smaller than the diffusion model.
This shift is especially helpful under strong source--target imbalance: ratio estimation only needs to separate domains, rather than model the full target distribution, and it admits flexible regularization that tends to stabilize training in data-scarce settings.
\newline \noindent \textbf{\textit{Relation to TLDM and generality of our formulation.}}
TLDM~\citep{ouyang2024transferlearningdiffusionmodels} derives a ratio-transfer rule only for \emph{continuous} data score-based diffusion in \(\mathbb R^d\), where guidance is expressed through Euclidean score gradients such as \(\nabla_{x_t}\log q_t(x_t)\). These gradients are not available for categorical tokens, and discrete diffusion is commonly trained by minimizing KL divergences between categorical posteriors rather than score-matching MSE. Our derivation therefore works directly with discrete forward and reverse kernels and the ELBO-style objective in \eqref{equ_mthd:theorm1_loss}. This formulation is architecture- and corruption-agnostic: it applies to a range of discrete forward processes (e.g., uniform, absorbing/masked, and general D3PM transitions) and does not depend on a particular parameterization of the denoiser in benefit to TLDM.
\newline \noindent \textbf{\textit{Continuous-Time GTL for Discrete Diffusion.}}
Because the argument is expressed at the level of the ELBO/KL objective, the same ratio-transfer principle carries over to continuous-time discrete diffusion whenever the training loss can be related back to an ELBO (e.g., \eqref{equ_bg:loss} and \eqref{equ_apdx:sedd_loss}). In Appendix (Theorem~\ref{Theorem:score}), we instantiate this transfer rule for continuous-time, score-based diffusion over discrete state spaces by reweighting the reverse-rate matrix with the learned ratio.

\paragraph{Sequence Formulation.}
We now apply Theorem~\ref{theorem1} to length-\(L\) token sequences.
As in prior work on discrete classifier guidance, we adopt a token-factorized reverse transition, which yields a practical sampler while keeping the denoiser unchanged. Concretely, we replace the classifier term \(p(y \mid \mathbf z_s^{(\ell)}, \mathbf z_t^{(1:L)})\) in \eqref{equ_bg:guidance} with a learned ratio model.
In the sequence setting, the ratio model is evaluated on a \emph{single-token intervention} of the current noisy sequence. Define \(\mathbf{z}_{t,s^\ell}^{(1:L)} \coloneqq [\,\mathbf{z}_t^{(1:\ell-1)},\,\mathbf{z}_s^{(\ell)},\,\mathbf{z}_t^{(\ell+1:L)}\,]\), i.e., the sequence obtained by replacing position \(\ell\) in \(\mathbf z_t^{(1:L)}\) by a candidate token \(\mathbf z_s^{(\ell)}\). With guidance strength \(\gamma \ge 0\), the resulting ratio-guided, factorized reverse transition is
\eqref{equ_mthd:extensiontosequence}.
\begin{align}
&q_{\theta, \phi}\!\left(\mathbf{z}_s^{(1:L)} \mid \mathbf{z}_t^{(1:L)}\right) = \prod_{\ell=1}^{L}
\frac{\Big[r_\phi\!\big(\mathbf{z}_{t,s^\ell}^{(1:L)}\big)\Big]^{\gamma}\;
p_\theta\!\left(\mathbf{z}_s^{(\ell)} \mid \mathbf{z}_t^{(1:L)}\right)}
{\displaystyle\sum_{\tilde{\mathbf{z}}_s^{(\ell)} \in \mathcal V}
\Big[r_\phi\!\big(\mathbf{z}_{t,\tilde s^\ell}^{(1:L)}\big)\Big]^{\gamma}\;
p_\theta\!\left(\tilde{\mathbf{z}}_s^{(\ell)} \mid \mathbf{z}_t^{(1:L)}\right)} .
\label{equ_mthd:extensiontosequence}
\end{align}
A naive implementation of \eqref{equ_mthd:extensiontosequence} is prohibitively expensive.
For each masked position \(\ell\), the denominator requires evaluating \(r_\phi(\mathbf z_{t,\tilde s^\ell}^{(1:L)})\) for all \(\tilde{\mathbf z}_s^{(\ell)} \in \mathcal V\), resulting in \(\mathcal O(L|\mathcal V|)\) ratio evaluations per denoising step.
This is manageable for small \(L\) or \(|\mathcal V|\) (e.g., our Markov-chain setting), but becomes prohibitive for text.

\begin{table}[t]
\centering
\footnotesize
\setlength{\tabcolsep}{3pt}
\renewcommand{\arraystretch}{1.1}
\begin{tabular}{lcc}
\toprule
\textbf{Method} & \textbf{Per-Step Complexity} & \textbf{Number of Steps ($T$)} \\
\midrule
Ancestral Sampling & $\mathcal{O}(|\mathcal V|\cdot L)$ & $T$ \\
+ Caching & $\mathcal{O}(|\mathcal V|\cdot L)$ & $T' \le \min(L,T)$ \\
+ Top-$n_{\text{ratio}}$ & $\mathcal{O}(n_{\text{ratio}}\cdot L)$ & $T' \le \min(L,T)$ \\
Planner Sampling \\
+ Top-$n_{\text{ratio}}$ & $\mathcal{O}(n_{\text{ratio}})$ & $L$ \\
\bottomrule
\end{tabular}
\caption{Computational cost per denoising step and total number of steps.}
\label{table_bg:complexity}
\end{table}
\begin{algorithm}[t]
\caption{Ratio-Guided Denoise Step $(\mathbf z_t, t, \Delta t)$.}
\label{alg:ratio_denoise}
\begin{algorithmic}[1]
\REQUIRE noisy sequence $\mathbf z_t \in \mathcal V^{L}$, denoiser $p_\theta$, ratio net $r_\phi$, guidance schedule $\gamma(\cdot)$, planner $\rho_\vartheta$
\STATE $\ell \gets \rho_\vartheta(\mathbf z_t)$ \hfill \COMMENT{Get position by planner}
\STATE $\log x_\theta \gets p_\theta(\mathbf z_t)$
\STATE $\mathcal N \gets \text{Top N tokens in }\log x_\theta[\ell,:]$
\FORALL{$v \in \mathcal N$}
  \STATE $\tilde{\mathbf z} \gets (\mathbf z_t[\ell]\gets v)$
  \STATE $\log r[\ell,v] \gets r_\phi(\tilde{\mathbf z},\sigma_t)$ \hfill \COMMENT{Ratio scores}
\ENDFOR
\STATE $\log q^{\text{guided}}[\ell,:] \gets \log x_\theta[\ell,:]+\gamma(t)\log r[\ell,:]$
\STATE $\log q^{\text{guided}}[\ell,\mathcal V \setminus \mathcal N] \gets -\infty$ \hfill \COMMENT{set non-Top-$N$ tokens $-\infty$}
\STATE $\pi \gets \operatorname{softmax}(\log q^{\text{guided}}[\ell,:])$
\STATE $\mathbf z_t^{(\ell)} \sim \operatorname{Cat}(\pi[\ell,:])$ \hfill \COMMENT{Sample at $\ell$}
\STATE $\mathbf z_s \gets \mathbf z_t$
\RETURN $\mathbf z_s$
\end{algorithmic}
\end{algorithm}

\subsection{Scalability via  Planner Sampling}
Our goal is to reduce ratio-network evaluations from \(\mathcal O(L|\mathcal V|)\) to a budget that is independent of \(|\mathcal V|\) and ideally weakly dependent on \(L\).
Table~\ref{table_bg:complexity} summarizes how caching, top-\(n_{\text{ratio}}\), and planner sampling progressively reduce this cost.
\paragraph{Ancestral Sampling} In the naive implementation, at all denoising steps, we compute logits from the denoiser and from the guidance network with $\mathcal{O}(|\mathcal V| \cdot L)$ complexity. 
\newline \noindent \textbf{\textit{Caching.}} In masked diffusion, the timestep is implicitly determined by the number of masked tokens, allowing us to remove explicit time conditioning from the networks~\citep{sahoo2024simpleeffectivemaskeddiffusion}.
As a result, if a denoising step does not modify the sequence, the corresponding logits can be reused. By caching logits across steps, the number of network evaluations becomes upper-bounded by the number of unmasking events. In the worst case, each position unmasks once, yielding at most \(L\) effective network calls. This optimization does not apply to uniform-noise diffusion, where token identities are mixed.
\noindent \textbf{\textit{Top-\(n_{\text{ratio}}\) pruning.}} Although caching reduces the number of steps, each step still requires evaluating the ratio network over the full vocabulary. To address this, we restrict ratio evaluations to a small candidate set. We assume that the denoiser distribution \(p_\theta(\cdot \mid \mathbf z_t)\) is sufficiently concentrated, so that meaningful deviations between the source and target distributions occur primarily among the top-\(n_{\text{ratio}}\) tokens. Formally, we assume that for these candidates the density ratio \(r(z)=q(z)/p(z)\) is bounded by a constant \(C<\infty\); no such bound is required outside this set, where \(p(z)\) is already negligible. Under this assumption, we evaluate the ratio term only for the top-\(n_{\text{ratio}}\) tokens proposed by the denoiser at each position. This produces position-specific candidate sets in the denominator of \eqref{equ_mthd:extensiontosequence} and reduces the cost of ratio evaluations to \(\mathcal O(n_{\text{ratio}}\,L)\), with \(n_{\text{ratio}}\ll|\mathcal V|\) in practice. \textbf{\textit{Mask-probability stabilizer.}} The interaction between top-\(n_{\text{ratio}}\) pruning and the guidance weight \(\gamma\) can distort probability mass, occasionally causing the diffusion process to collapse to fully masked sequences. To prevent this failure mode, we enforce a position-independent constraint that preserves the probability of remaining masked across source and target distributions. This induces a renormalized update over non-mask tokens and stabilizes the denoising dynamics. Details of this stabilization are provided in the appendix.

\paragraph{Planner Sampling} Even with top-\(n_{\text{ratio}}\) pruning, the overall complexity still scales linearly with \(L\).
To further reduce cost, we introduce a planner-based sampler that selects a single position to update at each step. The planner \(\rho_\vartheta\) is trained to predict the next denoising position in the masked sequence, replacing the stochastic Gillespie process with a deterministic \(\tau\)-leaping schedule. This strategy fixes the number of denoising steps to exactly \(L\), which is comparable to standard masked diffusion where typically \(T\approx L\), while substantially reducing guidance evaluations and wall-clock time.
Algorithm~\ref{alg:ratio_denoise} summarizes the resulting ratio-guided denoising step. Crucially, once the planner selects a position, ratio evaluations over the top-\(n_{\text{ratio}}\) candidates can be fully vectorized. As a result, each denoising step requires only a single forward pass through the denoiser, ratio network, and planner.
For planner training, we follow simplified strategies from \citet{peng2025pathplanningmaskeddiffusion, liu2025thinkgeneratediscretediffusion}.

\section{Experiments}
In this section, we present empirical results on the effectiveness of our method, Guided Transfer Learning for discrete diffusion models (GTL). We provide evidence across two applications showing that our method can successfully sample from the target distribution, in comparison to a vanilla diffusion model and standard finetuning. The first section presents a proof-of-concept on a synthetic dataset based on learning Markov chain data. The second subsection reports results on language modeling using arXiv abstracts.

\subsection{Synthetic Dataset}
\label{sec:markovchain}
\begin{figure*}[t]
\centering
\setlength{\tabcolsep}{2pt}
\renewcommand{\arraystretch}{1.0}
\begin{minipage}{0.7\textwidth}
\centering
\begin{tabular}{m{0.33\linewidth} m{0.65\linewidth}}
\begin{minipage}{\linewidth}\centering
  \includegraphics[width=\linewidth]{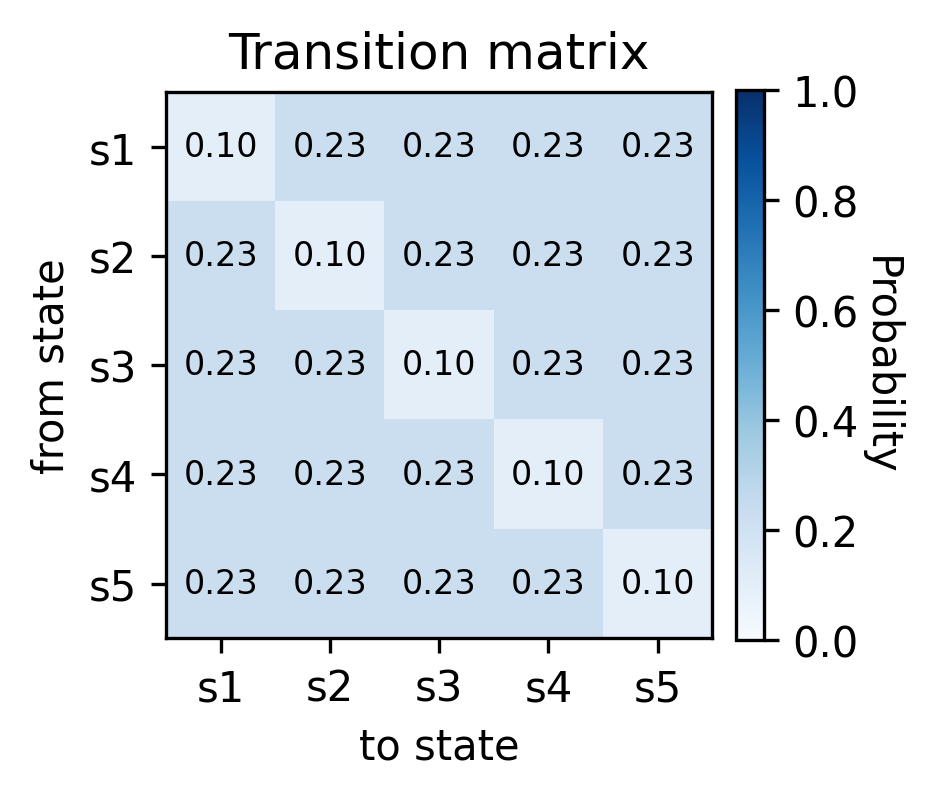}\\[-2pt]
  \small (a) Source\\[6pt]
  \includegraphics[width=\linewidth]{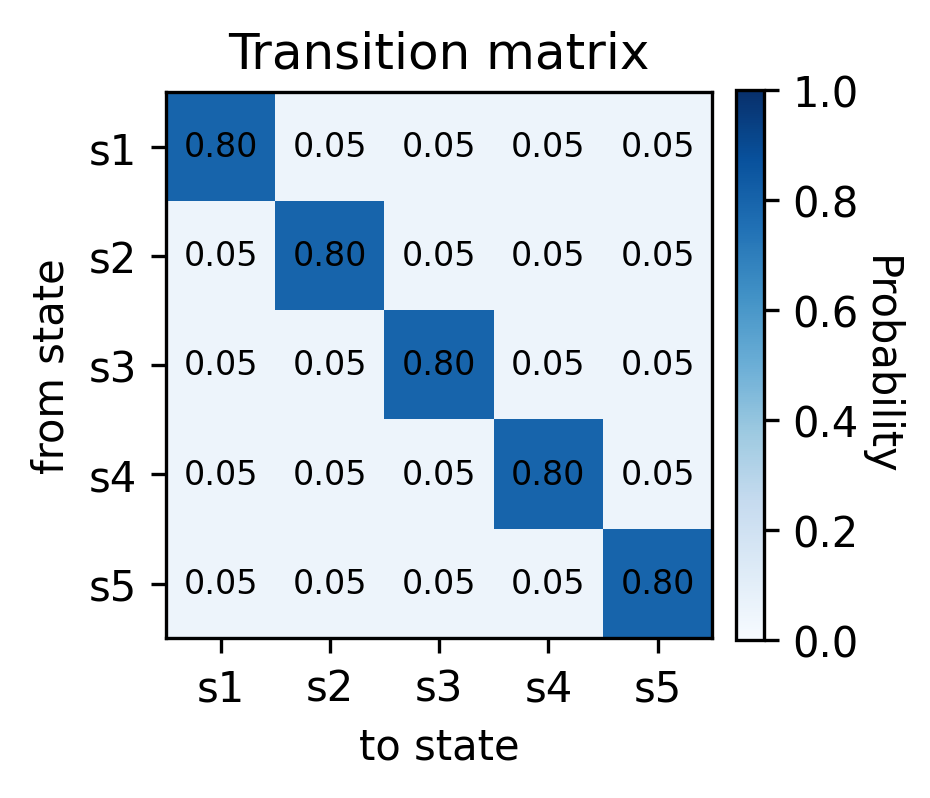}\\[-2pt]
  \small (b) Target
\end{minipage}
&
\begin{tabular}{M{0.10\linewidth} M{0.30\linewidth} M{0.30\linewidth} M{0.30\linewidth}}
$\boldsymbol{n}$ & \textbf{Vanilla Discrete Diffusion} & \textbf{Finetuned Discrete Diffusion} & \textbf{GTL} \\
\scriptsize 1000 &
\includegraphics[width=\linewidth]{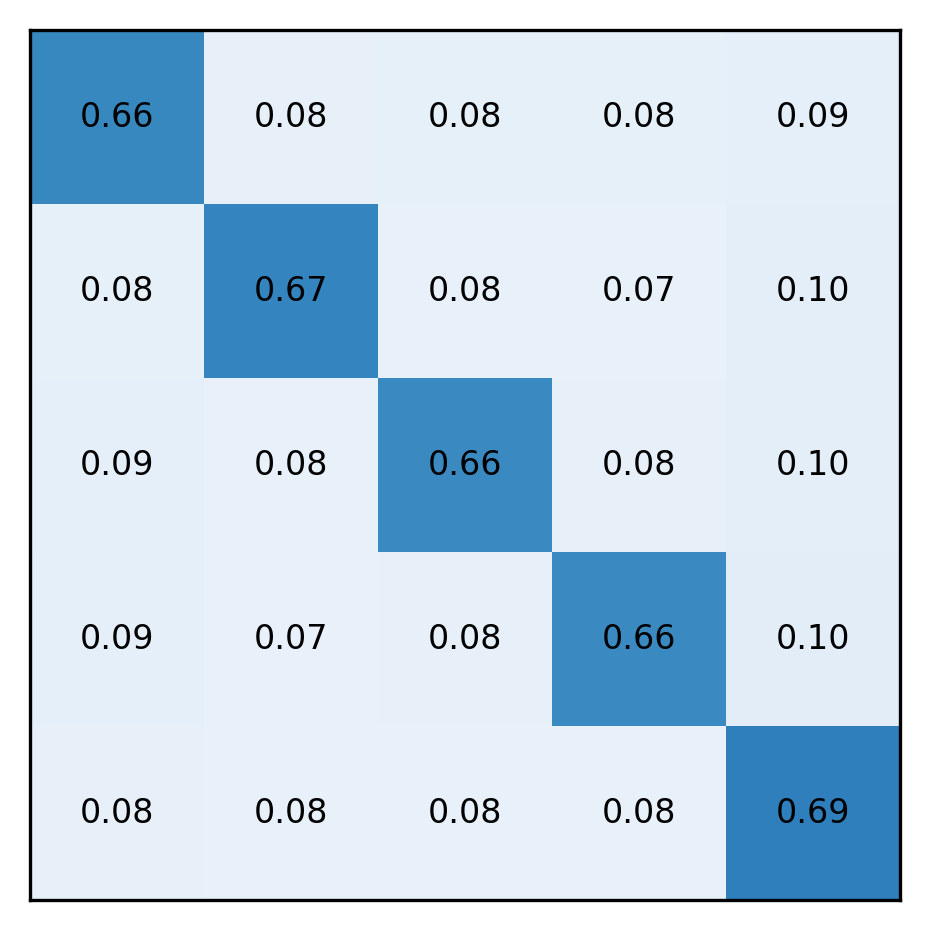} &
\includegraphics[width=\linewidth]{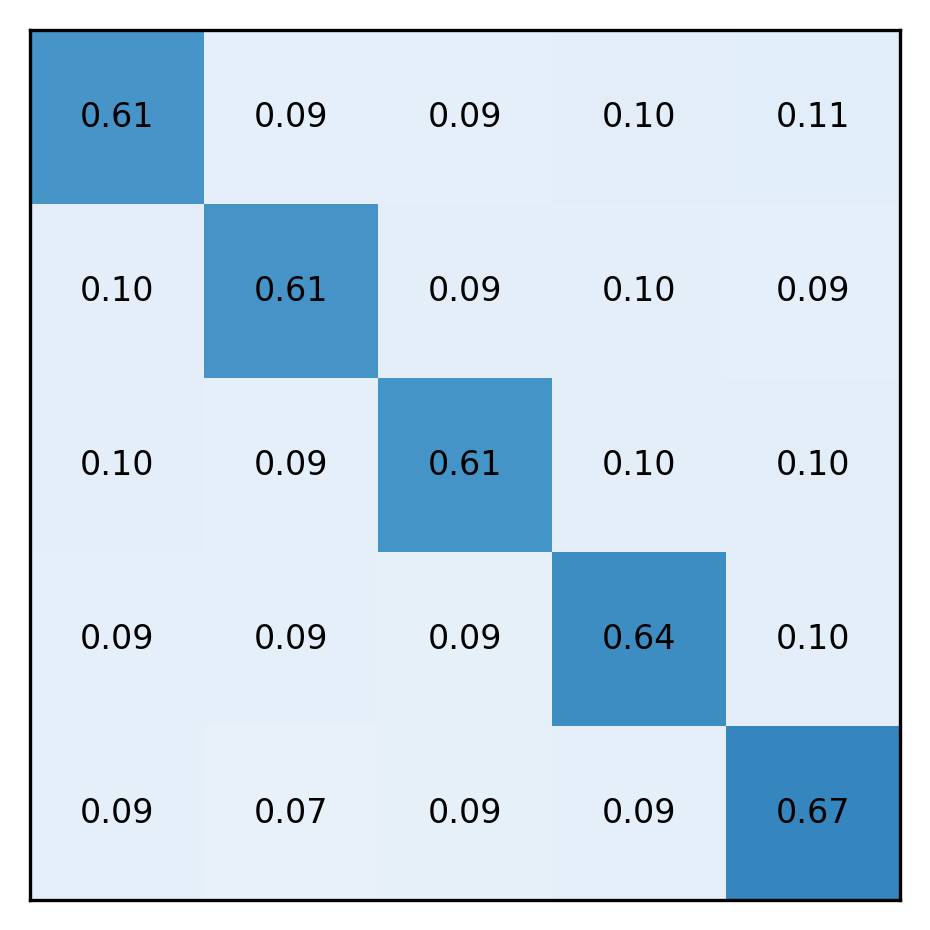} &
\includegraphics[width=\linewidth]{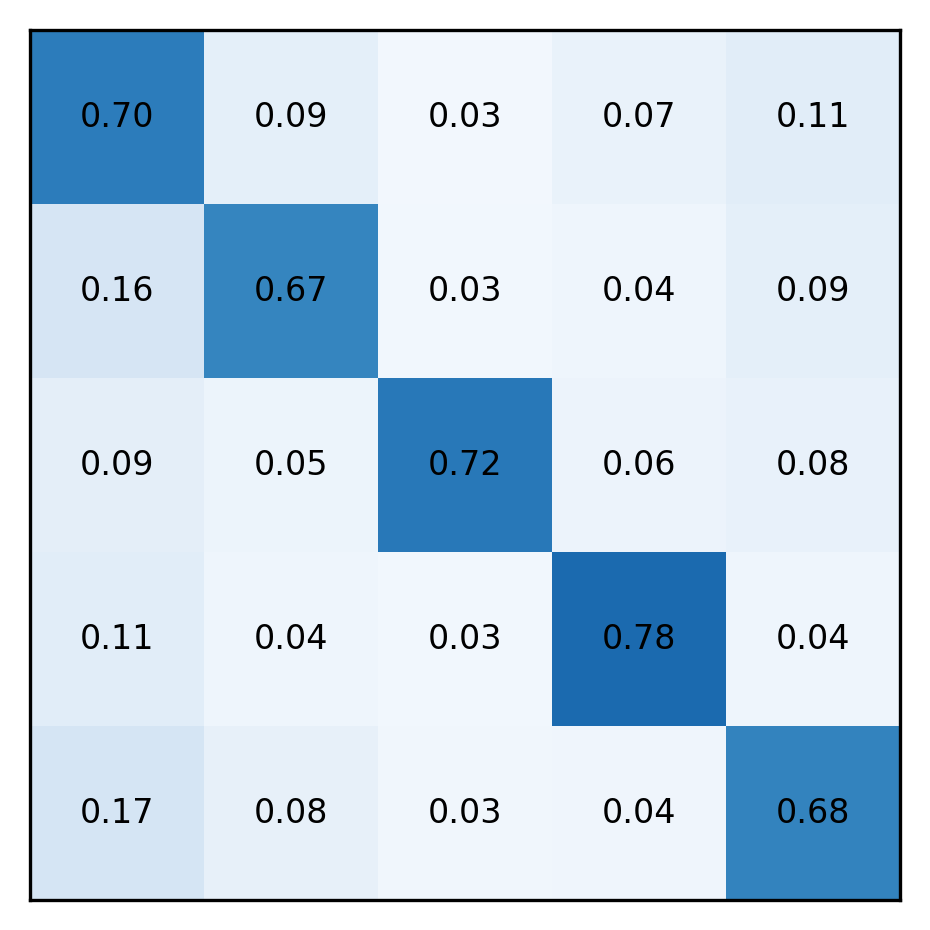} \\
\scriptsize 100 &
\includegraphics[width=\linewidth]{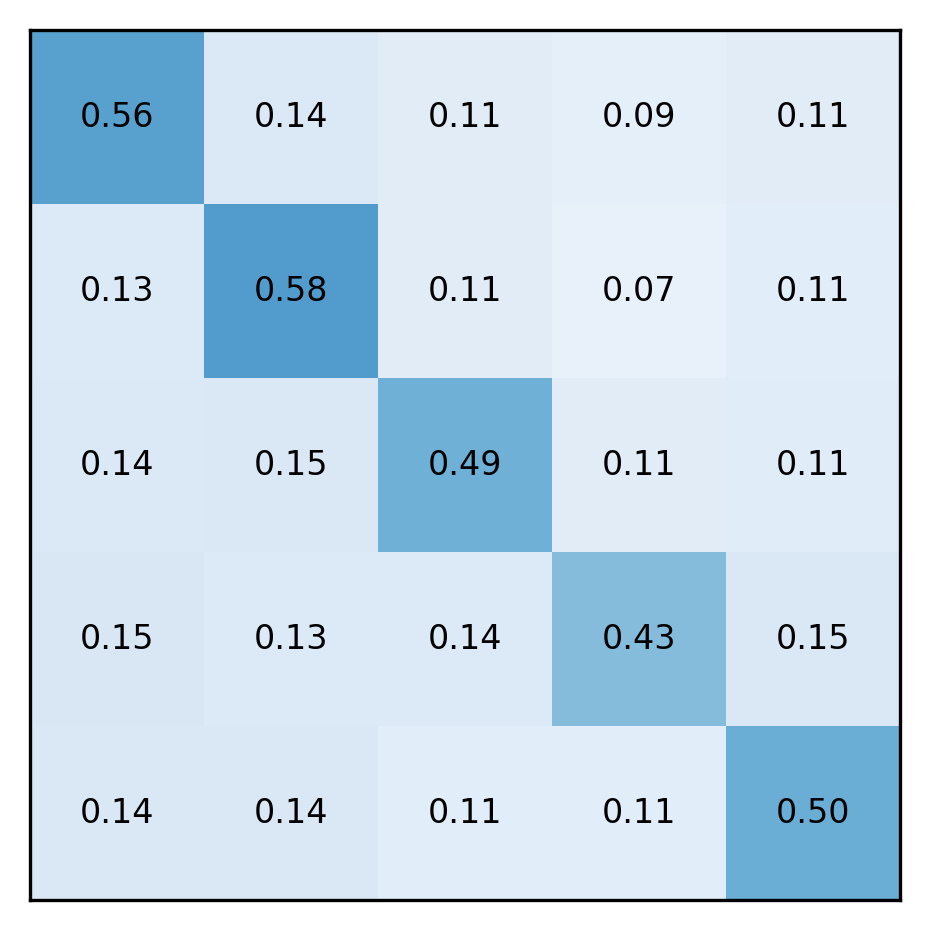} &
\includegraphics[width=\linewidth]{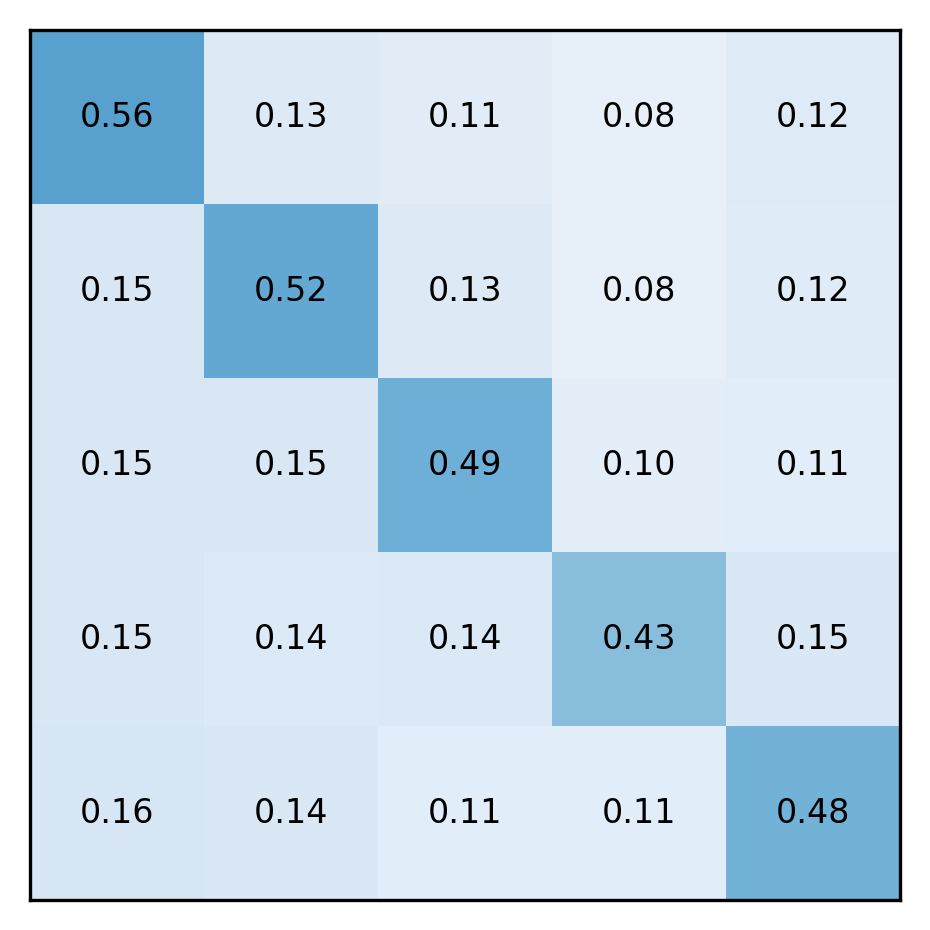} &
\includegraphics[width=\linewidth]{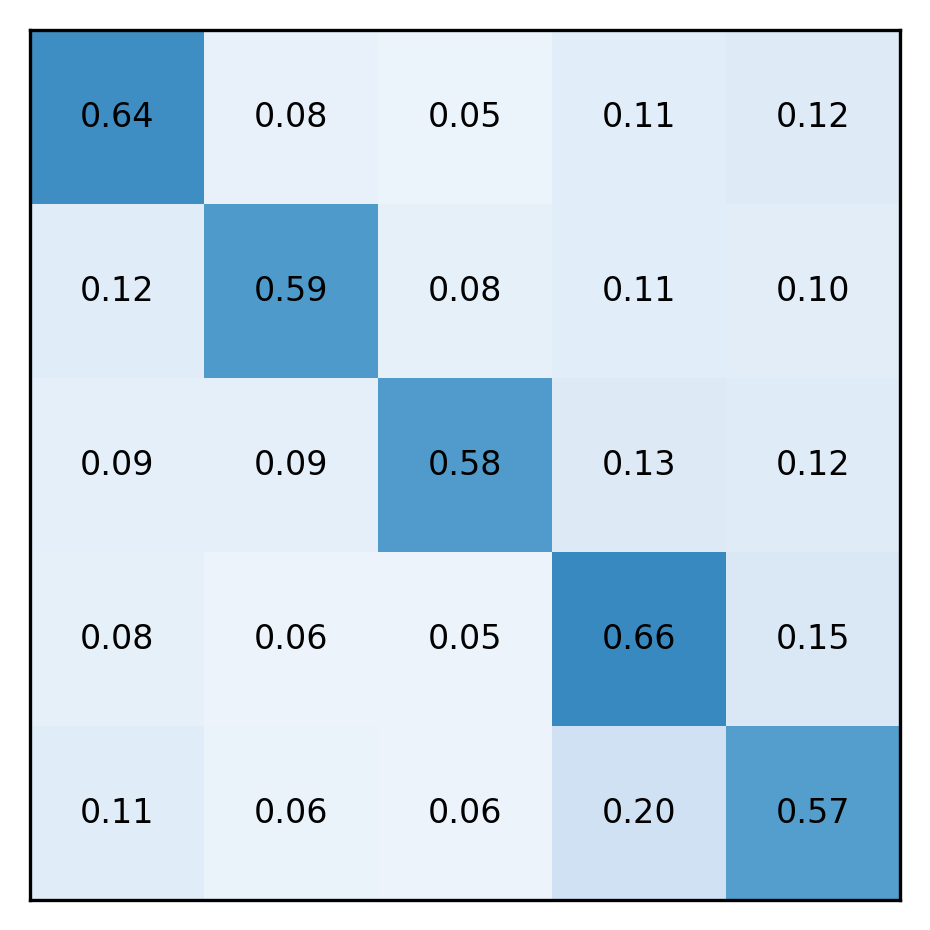} \\
\scriptsize 20 &
\includegraphics[width=\linewidth]{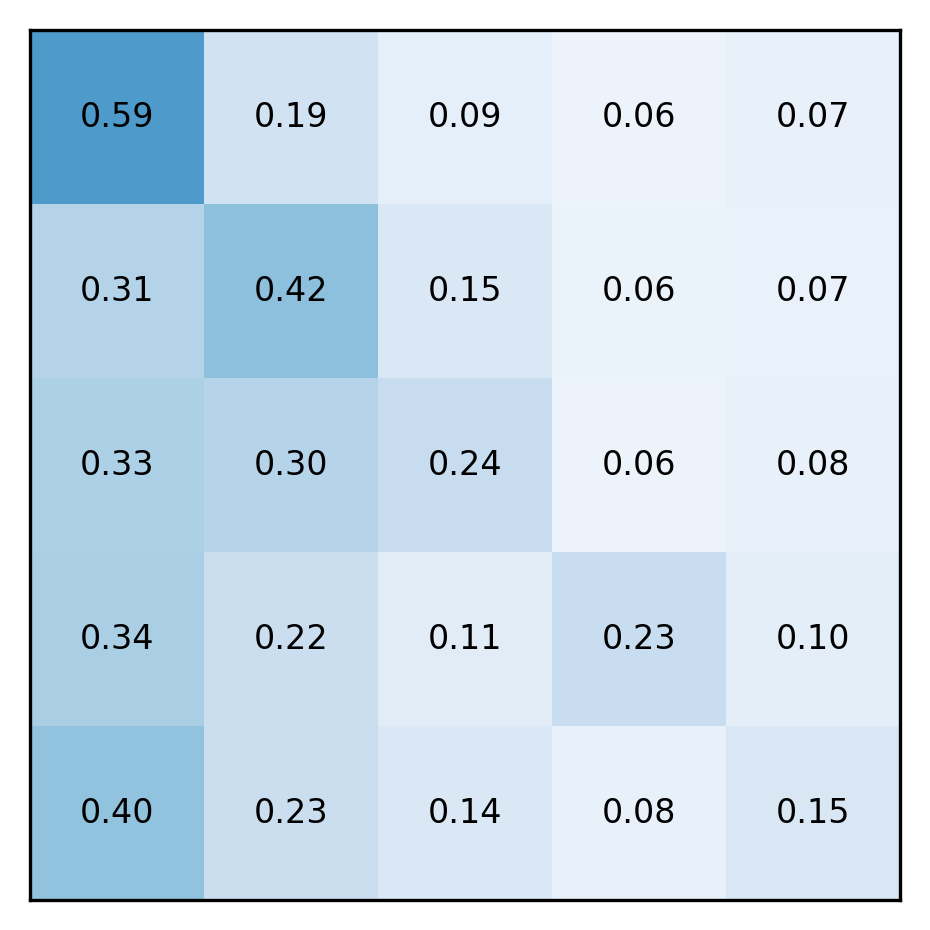} &
\includegraphics[width=\linewidth]{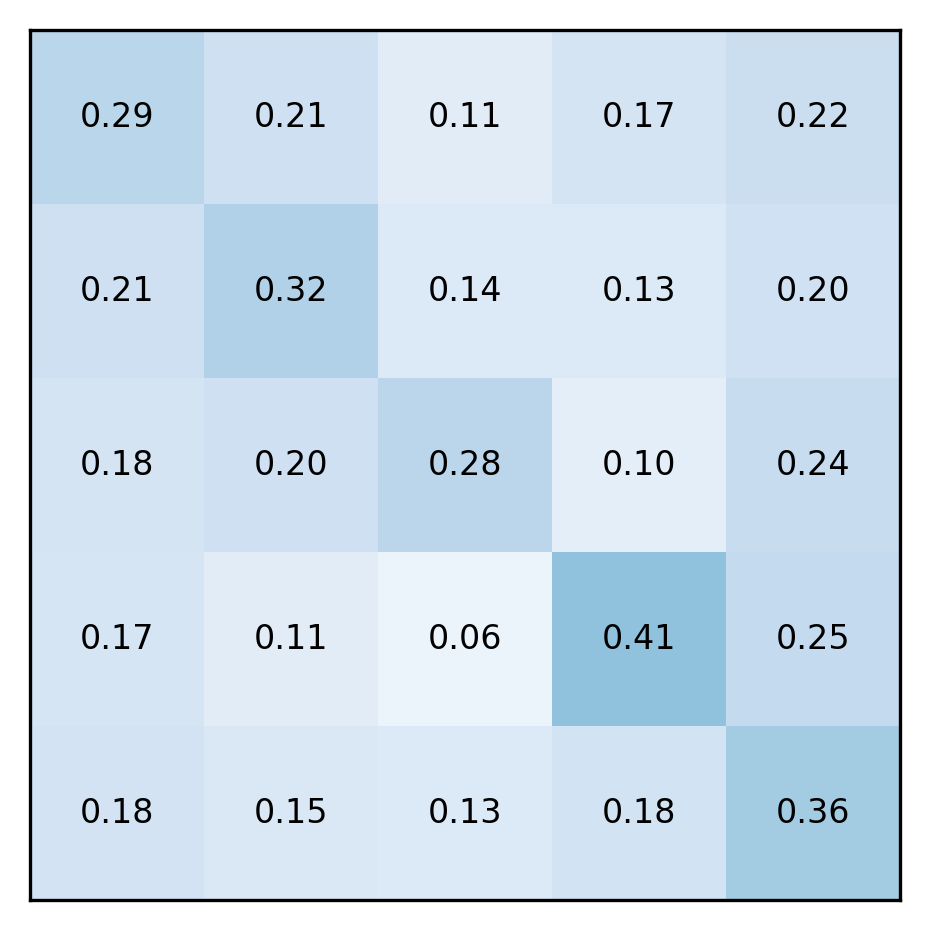} &
\includegraphics[width=\linewidth]{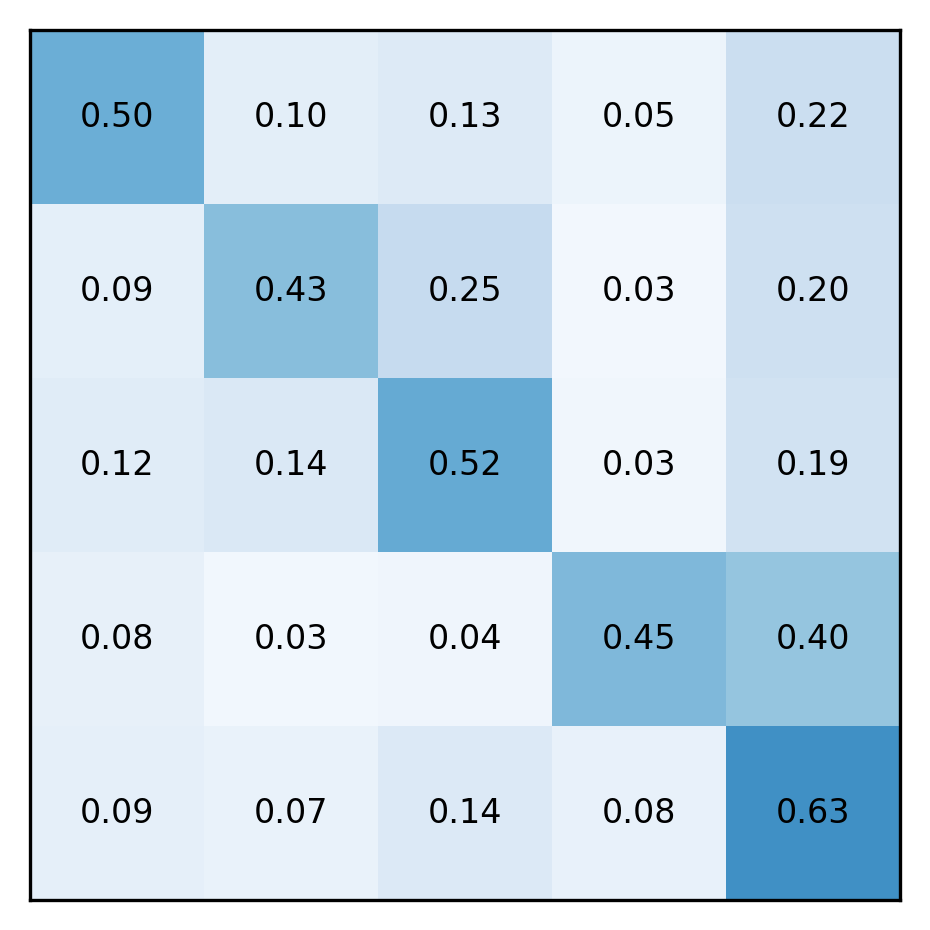} \\
\end{tabular}
\end{tabular}
\end{minipage}

\caption{True source and target (left). Estimated target transition matrices for varying number of target samples $n$ across three methods (right).}
\label{fig:source_plus_grid}
\end{figure*}
\noindent \textbf{Experimental Setup} We construct two Markov chains with distinct transition matrices. Let the probability state vector be $\mathbf{\pi}_i \in \Delta^{N-1}$ over a vocabulary of size $N=5$. The transition matrix of the target domain is $\mathbf{T} \in \mathbb{R}^{N \times N}$ and that of the source domain is $\mathbf{S} \in \mathbb{R}^{N \times N}$. We sample sequences of length $20$ from the target domain using the update rule $\mathbf{\pi}_i=\mathbf{T}\,\mathbf{\pi}_{i-1}$ with the initial state $\mathbf{\pi}_0$ set to the uniform vector. Sampling from the source domain proceeds in the same way. Implementation details can be found in the appendix \ref{sec:ImplementationMC}.
\newline \noindent \textbf{Results} We consider $m=10000$ source samples and $n \in \{1000,100,20\}$ target samples. We estimate the transition matrix by sampling $4096$ sequences, counting transitions between states, and normalizing by the total count. Figure~\ref{fig:source_plus_grid} presents an example where $\mathbf{S}$ has diagonal entries $0.1$ and $\mathbf{T}$ has diagonal entries $0.8$. We observe that the vanilla target diffusion model and the finetuned model both degrade in performance when fewer target samples are available.
\begin{table}[t]
\centering
\setlength{\tabcolsep}{3pt}   %
\footnotesize                 %
\begin{tabular}{@{}lccc@{}}   %
\toprule
$n$ & Vanilla (Target-only) & Finetuned & GTL \\
\midrule
1000 & $0.0476 \pm 0.0064$ & $0.0393 \pm 0.0029$ & $\mathbf{0.0377} \pm 0.0080$ \\
100  & $0.1938 \pm 0.0173$ & $0.1118 \pm 0.0271$ & $\mathbf{0.0989} \pm 0.0180$ \\
20   & $0.5842 \pm 0.0244$ & $0.4004 \pm 0.0983$ & $\mathbf{0.3621} \pm 0.0478$ \\
\bottomrule
\end{tabular}
\caption{Comparison of baseline methods and best guided results (mean $\pm$ standard error; $3$ seeds) across target sample sizes for $(\text{diag}_{\text{src}} = 0.1, \text{diag}_{\text{tgt}} = 0.8)$. Reported values are the average Kullback–Leibler divergence between true and estimated transition matrices; lower is better.}
\label{tab:synthetic_results}
\end{table}
Table~\ref{tab:synthetic_results} provides a direct comparison, obtained by running this example three times with different random seeds. We observe the following behaviors: (i) the diffusion model trained directly on the target dataset loses performance as the number of training samples decreases, and (ii) our method, compared to traditional finetuning, achieves better results even with fewer trainable parameters, while showing a similar trend in performance across different numbers of training samples.
\newcommand{\figScale}{0.66}

\subsection{Language Modeling}
\label{sec:exp_languagemodeling}
\begin{figure*}[ht]
    \centering
    \includegraphics[width=\textwidth]{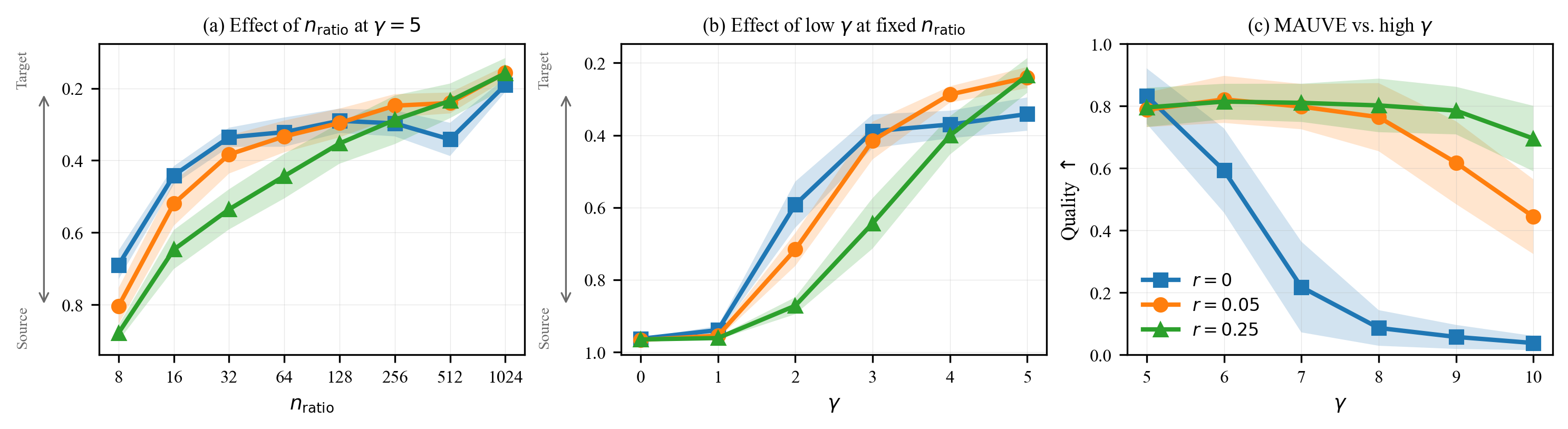}
    \caption{Sensitivity of GTL to the guidance weight $\gamma$ and the candidate budget $n_{\text{ratio}}$ on Physics validation data.
    The source domain is $\mathcal{S}=\mathrm{CS}\cup\mathrm{Math}\cup\mathcal{P}_r$ and the target domain is $\mathcal{T}=\mathcal{P}\setminus\mathcal{P}_r$, where $\mathcal{P}_r$ is a random $r$-subset of Physics.
    We use $r\in\{0,0.05,0.25\}$, where $r$ denotes the fraction of Physics included in the source.
    For each setting, we train three independent source models and three independent ratio models.
    Lines show means and shaded bands show $\pm$ standard error over $3\times3$ train$\times$sample seeds. Increasing $n_{\text{ratio}}$ improves target alignment, while overly strong guidance reduces quality most strongly when $r=0$.}
    \label{fig:behavior}
\end{figure*}

In this section, we apply our method to text data. We use the arXiv abstracts dataset \citep{arxiv_org_submitters_2024}, which contains 1.7 million arXiv articles with titles, authors, categories, and abstracts. We focus on Computer Science, Mathematics, and Physics. For data preparation, we concatenate all text and split it into segments of 512 tokens using the \texttt{bert-base-uncased} tokenizer with vocabulary size $N=30{,}522$. This results in $285{,}946$ samples in Computer Science, $176{,}831$ in Mathematics, and $79{,}631$ in Physics. For each category we randomly select 10\% of the samples for validation. We ensure that the three category subsets are disjoint. 
\newline \noindent We run two types of experiments. First, we study the behavior of the two key parameters $\gamma$ and $n_{\text{ratio}}$. Second, we compare three settings on the target domain. The first is a vanilla discrete diffusion model trained on the target data. The second is a finetuned version of the source model. The third is our guided transfer method.
\newline \noindent \textbf{Implementation and Evaluations}
We train the source diffusion model for 100{,}000 gradient steps using the DiT architecture \citep{peebles2023scalablediffusionmodelstransformers} with 59.8M trainable parameters. A log–linear noise scheduler is used for both training and sampling. The finetuned model on the target domain is trained for 10{,}000 gradient steps. For our guided method, we initialize a ratio model with 4.1M parameters and train it for 5{,}000 gradient steps. Additional training details appear in the supplementary materials.
For each method we generate 128 sequences of length 512. We evaluate quality with MAUVE \citep{pillutla2021mauvemeasuringgapneural}, which captures the tradeoff between perplexity and diversity. Although generative perplexity (Gen. PPL) is often reported, we observed, consistent with prior work \citep{wang2025remaskingdiscretediffusionmodels}, that Gen. PPL can remain low even when the text has reduced utility, such as repeated words. In those cases, MAUVE is more sensitive and provides a clearer measure of quality. We also evaluate the domain of each sampled sequence using an independently trained classifier. Scores near zero indicate the target domain and scores near one indicate the source domain. The classifier follows the setup used for the ratio model in the supplementary materials with label smoothing set to $0.05$.
\subsubsection{Behavior of $\gamma$ and $n_{\text{ratio}}$}
We study the sensitivity of GTL to the guidance weight $\gamma$ and the candidate budget $n_{\text{ratio}}$. For each configuration, we train three independent source models and three ratio models using different random seeds. The source domain is defined as Computer Science $\cup$ Mathematics plus an $r$-fraction of Physics abstracts, and the target domain is the remaining $(1-r)$ fraction of Physics. 
Formally, for $r \in [0,1]$, let $\mathcal{P}_r$ be a random $r$-subset of Physics (sampled without replacement). We set $\mathcal{S} = \text{CS} \cup \text{Math} \cup \mathcal{P}_r$ and $\mathcal{T} = \mathcal{P} \setminus \mathcal{P}_r$. We set $r \in \{0, 0.05,0.25\}$, thus in the case of $r=0$ we have two distinct datasets. This setup simulates the common scenario where pretraining large corpora already includes a small portion of target-domain text.
In Figure~\ref{fig:behavior}, the behavior of these parameters is illustrated. Intuitively, smaller candidate budgets $n_{\text{ratio}}$ introduce higher sampling error, whereas increasing $n_{\text{ratio}}$ monotonically reduces the domain-classifier score, yielding more target-like samples, as shown in panel (a). In Figure~\ref{fig:behavior}(b), we observe that the guidance score is nontrivial: a moderate $\gamma \approx 3\!-\!5$ consistently pushes the representations toward the target across all $r$. 
Panel (c) highlights two key effects: (i) excessive guidance can degrade performance when there is no overlap between source and target data, and (ii) when a small portion of target text is included in the source domain ($r = 0.05, 0.25$), MAUVE remains high for larger $\gamma$ values and decays more smoothly. This indicates that a higher $r$ improves robustness. Overall, the results suggest that $n_{\text{ratio}} \approx 256\!-\!512$ and $\gamma \approx 4\!-\!6$ achieve a favorable balance, providing strong target alignment with minimal quality loss.

\subsubsection{Method Comparison and Sensitivity to Data Fraction}
In this section, we define the union of Computer Science and Mathematics abstracts as the source domain and Physics abstracts as the target domain. We compare three methods: a fine-tuned discrete diffusion model trained on the target domain after pretraining on the source domain, a vanilla discrete diffusion model trained directly on the target domain, and our Guided Transfer Learning (GTL) method with the ratio estimator trained across target-data fractions of 100\%, 30\%, 5\%, and 1\%. Each method is trained with three random seeds, and sampling is performed with three seeds as well (nine runs per cell). Furthermore, each seed corresponds to a distinct subset of the target data. Figure~\ref {fig:gtl_overview_and_results} reports the behavior of the models across different fractions of the target training data. The observed trends align with those in Table~\ref{tab:synthetic_results}. In the smallest case, we only use 796 samples of Physics abstracts. For finetuned and vanilla discrete diffusion, the number of epochs over the training set increases up to 400 when the fraction is smallest. The ratio model decreases its MAUVE score with decreasing fractions, but less dramatically than the other methods. Thus, our method demonstrates greater stability under limited target-domain data. In contrast, the finetuned and vanilla discrete diffusion models show a stepwise drop in performance as the fraction decreases. Even in the full-target-data setting, GTL outperforms the baselines because there is still a large imbalance between the number of source-domain and target-domain samples. The GTL remains relatively stable, except in the 1\% case, where we also observe a sharp decline. %
\section{Conclusion}
Adapting discrete diffusion models to target domains with limited data is often costly. To address this, we propose a guided transfer learning approach for discrete diffusion models that introduces a small, learned ratio network, allowing the training of significantly fewer parameters. Furthermore, we present a guidance-based sampling algorithm that enables generation with larger vocabularies and longer sequences, making guided language modeling with discrete diffusion models feasible. We demonstrate the effectiveness of our method on synthetic Markov chains and arXiv abstracts (language modeling), particularly when the target dataset is small relative to the source.
\noindent \newline\textbf{Limitations}
Overall, the proposed method is promising and applicable to a wide range of tasks involving domain shifts. However, it requires access to both source and target domains. Future work could explore conditional transfer for prompt-based or attribute-controlled generation. It would also be interesting to evaluate the learned ratio on quality-filtered tasks, such as reducing toxicity in language modeling.

\bibliography{iclr2026_conference}
\bibliographystyle{iclr2026_conference}

\appendix
\section{Appendix}

\section{Related Work}
\paragraph{Discrete Diffusion Models}
Discrete diffusion models (DDMs) have become a standard choice for generative modeling over categorical sequences, including biological sequences (DNA, RNA, proteins) and text~\citep{Sarkar2024, yu2025discretediffusionlargelanguage}. Compared to approaches that add continuous noise to embeddings or latents, fully discrete formulations avoid the projection step back to tokens, which can introduce an embedding-to-token mismatch~\citep{li2022diffusionlmimprovescontrollabletext, lovelace2023latentdiffusionlanguagegeneration}. Two families of discrete diffusion methods are especially common. The first is discrete-time diffusion in the D3PM framework, where a Markov corruption process is defined by transition matrices and a denoiser is trained to invert the forward kernels; simplified masked variants such as MLDM and MD4 fall into this category~\citep{sahoo2024simpleeffectivemaskeddiffusion, shi2025simplifiedgeneralizedmaskeddiffusion, austin2023structureddenoisingdiffusionmodels}. The second is continuous-time discrete diffusion, which parameterizes a discrete score (a log-ratio between neighboring-time marginals) and induces a continuous-time Markov chain (CTMC) for generation~\citep{lou2024discretediffusionmodelingestimating, meng2023concretescorematchinggeneralized, campbell2022continuoustimeframeworkdiscrete}.
\paragraph{Guided Diffusion Models}
Guidance is a widely used mechanism for steering diffusion sampling in both continuous and discrete domains. In discrete diffusion, classifier-free guidance typically combines conditional and unconditional predictions at inference time, requiring two forward passes of a single model trained on a mixture of conditional and unconditional examples~\citep{huang2025ctrldiffboostinglargediffusion}. Classifier-based guidance is closer to our setting: it introduces an auxiliary classifier $p(y \mid x_t, t)$ and reweights the reverse transition so that samples are biased toward the desired attribute y, while the denoiser itself is trained on the full corpus~\citep{schiff2025simpleguidancemechanismsdiscrete, nisonoff2025unlockingguidancediscretestatespace}. A direct implementation scores token substitutions across all positions and vocabulary items, requiring $\mathcal{O}(L|\mathcal{V}|)$ guidance evaluations per step, which is infeasible for long sequences and large vocabularies. Because each evaluation corresponds to a distinct single-token intervention on the full sequence, this cost is not removed by straightforward parallelization. To reduce this cost, \citep{nisonoff2025unlockingguidancediscretestatespace} proposes a first-order Taylor approximation that replaces many evaluations with a forward and backward pass. Related alternatives adjust sampling using inference-time optimization: TESS2~\citep{tae2025tess2largescalegeneralist} performs gradient ascent on a reward with respect to the model logits, and energy-based diffusion language models define an energy over predicted clean sequences and use it to reweight denoising updates to reduce train–sample mismatch~\citep{xu2025energybaseddiffusionlanguagemodels}.

\paragraph{Transfer Learning}
Fine-tuning and transfer learning share the goal of adapting a model to a new task or domain. Fine-tuning is computationally expensive because it updates the pretrained denoiser’s weights on new data. Prior work on continuous diffusion models explores reinforcement learning~\citep{fan2023dpokreinforcementlearningfinetuning, clark2024directlyfinetuningdiffusionmodels} and adapter-based fine-tuning~\citep{moon2022finetuning, xie2023difffitunlockingtransferabilitylarge}. In contrast, transfer learning for diffusion remains comparatively underexplored. \emph{Transfer Learning for Diffusion Models} (TLDM) (\citet{ouyang2024transferlearningdiffusionmodels}) proposes a ratio-estimation technique for continuous score-based models,but it relies on continuous scores and is evaluated in low-dimensional settings. Our work closes the gap for discrete diffusion by deriving ratio-guided reverse transitions in both discrete-time, continuous-time and score based formulations for high dimansion application.

\section{Proof of Theorem 1}
\label{sec_apx:proof}
For the proof, we use the formulation of importance reweighting as described in the following lemma.
\begin{lemma}[Importance-weighted reformulation]\label{lemma:importance_reformulation}
Fix a timestep \(i\) and define \(r(\mathbf x_0)=q(\mathbf x_0)/p(\mathbf x_0)\).
If the source and target share the same forward process, then the loss function can be rewritten as: 
\begin{align}
\mathcal L_i(\psi)&=
\mathbb E_{q}\bigl[
  D_{\mathrm{KL}}\!\bigl(
     q(\mathbf z_{s(i)}\mid\mathbf z_{t(i)},\mathbf x_0)
     \,\|\,q_\psi(\mathbf z_{s(i)}\mid\mathbf z_{t(i)})
  \bigr)
\bigr]
\\&=
\mathbb E_{\mathbf x_0\sim p\;
 z_{t(i)}\sim p(\mathbf z_{t(i)}\mid\mathbf x_0)}
\Bigl[r(\mathbf x_0)
  D_{\mathrm{KL}}\!\bigl(
    p(\mathbf z_{s(i)}\mid\mathbf z_{t(i)},\mathbf x_0)
    \,\|\,q_\psi(\mathbf z_{s(i)}\mid\mathbf z_{t(i)})
  \bigr)\,
\Bigr].
\end{align}
\end{lemma}
\paragraph{Proof of Lemma \ref{lemma:importance_reformulation}}
\begin{align}
\mathbb{E}_{q}\!\left[\mathcal{L}_{\text{diffusion}}\right]
&=  
\sum_{i=1}^{T}
\mathbb{E}_{q}\!\Bigl[D_{\mathrm{KL}}\!\bigl(
q(\mathbf z_{s(i)}\mid\mathbf z_{t(i)},\mathbf x_0)\,\big\|\,q_\psi(\mathbf z_{s(i)}\mid\mathbf z_{t(i)})
\bigr)\Bigr] 
\\&= 
\sum_{i=1}^{T}
\sum_{\mathbf x_0}\sum_{\mathbf z_{t(i)}}
q(\mathbf x_0,\mathbf z_{t(i)})
D_{\mathrm{KL}}\!\bigl(
q(\mathbf z_{s(i)}\mid\mathbf z_{t(i)},\mathbf x_0)\,\big\|\,q_\psi(\mathbf z_{s(i)}\mid\mathbf z_{t(i)})
\bigr) 
\\&= 
\sum_{i=1}^{T}
\sum_{\mathbf x_0}\sum_{\mathbf z_{t(i)}}
q(\mathbf x_0)\,q(\mathbf z_{t(i)}\mid\mathbf x_0)
D_{\mathrm{KL}}\!\bigl(
q(\mathbf z_{s(i)}\mid\mathbf z_{t(i)},\mathbf x_0)\,\big\|\,q_\psi(\mathbf z_{s(i)}\mid\mathbf z_{t(i)}) \bigr)
\\&= 
\sum_{i=1}^{T} \sum_{\mathbf x_0}\sum_{\mathbf z_{t(i)}} {\frac{q(\mathbf x_0)}{p(\mathbf x_0)}} p(\mathbf x_0)\,p(\mathbf z_{t(i)}\mid\mathbf x_0) D_{\mathrm{KL}}\!\bigl( q(\mathbf z_{s(i)}\mid\mathbf z_{t(i)},\mathbf x_0)\,\big\|\,q_\psi(\mathbf   z_{s(i)}\mid\mathbf z_{t(i)}) \bigr) \label{proof1:lemma_p} 
\\&=
\sum_{i=1}^{T} \mathbb E_{\mathbf x_0\sim p}\;\mathbb E_{\mathbf z_{t(i)}\sim p(\mathbf z_{t(i)}\mid\mathbf x_0)}\Bigl[ \frac{q(\mathbf x_0)}{p(\mathbf x_0)}D_{\mathrm{KL}}\!\bigl(
p(\mathbf z_{s(i)}\mid\mathbf z_{t(i)},\mathbf x_0) \,\|\,q_\psi(\mathbf z_{s(i)}\mid\mathbf z_{t(i)}) \bigr)\, \Bigr] \label{proof1:lemma_q} 
\end{align}
To conclude Lemma \ref{lemma:importance_reformulation}, we leverage the fact that the source and target domain share the same forward kernel in Equation \ref{proof1:lemma_p} to \ref{proof1:lemma_q} 
$ q(\mathbf z_{s(i)} \mid \mathbf z_{t(i)}, \mathbf x_0) = \frac{q(\mathbf z_{t(i)} \mid \mathbf z_{s(i)})\cdot q (\mathbf z_{s(i)} \mid \mathbf x_{0})}{q (\mathbf z_{t(i)} \mid \mathbf x_{0})}  = 
p(\mathbf z_{s(i)} \mid \mathbf z_{t(i)}, \mathbf x_0) $. 
\newline
\noindent
Now, we use Lemma \ref{lemma:importance_reformulation}:
\begin{align}
\mathcal L_i(\psi) &=
\mathbb E_{\mathbf x_0\sim p}\;
\mathbb E_{\mathbf z_{t(i)}\sim p(\mathbf z_{t(i)}\mid\mathbf x_0)}
\Bigl[\underbrace{\frac{q(\mathbf x_0)}{p(\mathbf x_0)}}_{r(\mathbf x_0)}
D_{\mathrm{KL}}\!\bigl(
p(\mathbf z_{s(i)}\mid\mathbf z_{t(i)},\mathbf x_0)
\,\|\,q_\psi(\mathbf z_{s(i)}\mid\mathbf z_{t(i)})
\bigr)\, \Bigr]
\\& = 
\sum_{\mathbf{x}_0} \sum_{\mathbf{z}_{t(i)}} \underbrace{p(\mathbf z_{t(i)}\mid\mathbf x_0) \; p(\mathbf x_0)}_{p(\mathbf x_{0}\mid\mathbf z_{t(i)}) \; p(\mathbf z_{t(i)})}  r(\mathbf x_0) D_{\mathrm{KL}}\bigl(p(\mathbf z_{s(i)}\mid\mathbf z_{t(i)},\mathbf x_0)\;\|\;q_\psi(\mathbf z_{s(i)}\mid\mathbf z_{t(i)}) \bigr) 
\\& = 
  \sum_{\mathbf{z}_{t(i)}} p(\mathbf z_{t(i)}) \sum_{\mathbf{x}_0}  \underbrace{p(\mathbf x_{0}\mid\mathbf z_{t(i)}) r(\mathbf{x}_0)}_{\alpha(\mathbf{x}_0,\mathbf{z}_{t(i)})} D_{\mathrm{KL}}\bigl(p(\mathbf z_{s(i)}\mid\mathbf z_{t(i)},\mathbf x_0)\;\|\;q_\psi(\mathbf z_{s(i)}\mid\mathbf z_{t(i)}) \bigr)
\\& = 
\sum_{\mathbf{z}_{t(i)}} p(\mathbf z_{t(i)}) \sum_{\mathbf{x}_0}  \alpha(\mathbf{x}_0,\mathbf{z}_{t(i)}) \; D_{\mathrm{KL}}\bigl(p(\mathbf z_{s(i)}\mid\mathbf z_{t(i)},\mathbf x_0)\;\|\;q_\psi(\mathbf z_{s(i)}\mid\mathbf z_{t(i)}) \bigr)
\end{align}
Now, we take the derivative in regard to $\psi$:
\begin{align}
    &\qquad \qquad 0=
    \frac{\partial}{\partial \psi} \mathcal L_i(\psi) 
    \\\Longrightarrow & \qquad\qquad 0=
    \frac{\partial}{\partial \psi} \sum_{\mathbf{x}_0}  \alpha(\mathbf{x}_0,\mathbf{z}_{t(i)}) \; D_{\mathrm{KL}}\bigl(p(\mathbf z_{s(i)}\mid\mathbf z_{t(i)},\mathbf x_0)\;\|\;q_\psi(\mathbf z_{s(i)}\mid\mathbf z_{t(i)}) \bigr)
\end{align}
To solve this, we set up a Lagrangian equation by using the constraint $\sum_{\mathbf z_{s(i)}}q_\psi(\mathbf z_{s(i)}\mid\mathbf z_{t(i)})=1$
\begin{align}
 \mathcal L(\psi,\lambda)\;=\; &\sum_{\mathbf{x}_0}  \alpha(\mathbf{x}_0,\mathbf{z}_{t(i)}) \sum_{\mathbf z_{s(i)}} p(\mathbf z_{s(i)} \mid \mathbf z_{t(i)}, \mathbf x_0) \bigl[ \log p(\mathbf z_{s(i)} \mid \mathbf z_{t(i)}, \mathbf x_0) - \log q_\psi(\mathbf z_{s(i)} \mid \mathbf z_{t(i)}) \bigr] \nonumber\\&+\;\lambda\!\Bigl(\sum_{\mathbf z_{s(i)}}q_\psi(\mathbf z_{s(i)}\mid\mathbf z_{t(i)})-1\Bigr)
\end{align}
Stationary condition for each \(\mathbf z_{s(i)}\):
\begin{align}
    &\frac{\partial\mathcal L}{\partial \psi} = -\frac{\sum_{\mathbf x_0}\alpha(\mathbf x_0,\mathbf z_{t(i)})\,p(\mathbf z_{s(i)}\mid\mathbf z_{t(i)},\mathbf x_0)} {q_\psi(\mathbf z_{s(i)}\mid\mathbf z_{t(i)})} \;+\;\lambda=0 \quad 
    \\&
    \Longrightarrow\quad {q_\psi(\mathbf z_{s(i)}\mid\mathbf z_{t(i)})= \frac{\sum_{\mathbf x_0}\alpha(\mathbf x_0,\mathbf z_{t(i)})\,p(\mathbf z_{s(i)}\mid\mathbf z_{t(i)},\mathbf x_0)}
     {\lambda}}
\end{align}
Now we determine \(\lambda\) from the constraint
\begin{align}
    1 &=\sum_{\mathbf z_{s(i)}}q_\psi(\mathbf z_{s(i)}\mid\mathbf z_{t(i)})
    =\sum_{\mathbf z_{s(i)}}
    \frac{1}{\lambda} \sum_{\mathbf x_0}\alpha(\mathbf x_0,\mathbf z_{t(i)})\,p(\mathbf z_{s(i)}\mid\mathbf z_{t(i)},\mathbf x_0)
    \\
    \;\Longleftrightarrow\qquad \lambda &=\sum_{\mathbf z_{s(i)}}
     \sum_{\mathbf x_0}\alpha(\mathbf x_0,\mathbf z_{t(i)})\,p(\mathbf z_{s(i)}\mid\mathbf z_{t(i)},\mathbf x_0)
\end{align}
Putting all together:
\begin{align}
q_{\psi^\star}(\mathbf z_{s(i)}\mid\mathbf z_{t(i)}) &=
\frac{\displaystyle\sum_{\mathbf x_0}\alpha(\mathbf x_0,\mathbf z_{t(i)})\,p(\mathbf z_{s(i)}\mid\mathbf z_{t(i)},\mathbf x_0)}{\displaystyle\sum_{\mathbf{\tilde{z}}_{s(i)}}\displaystyle\sum_{\mathbf x_0}\alpha(\mathbf x_0,\mathbf z_{t(i)}) p(\mathbf{\tilde{z}}_{s(i)}\mid\mathbf z_{t(i)},\mathbf x_0)}  \label{equ_wk10:sol_alpha}
\\&=
\frac{p(\mathbf z_{s(i)} \mid \mathbf z_{t(i)}) \frac{q(\mathbf z_{s(i)})}{p(\mathbf z_{s(i)})}}{\displaystyle\sum_{\mathbf{\tilde{z}}_{s(i)}}p(\mathbf{\tilde{z}}_{s(i)}\mid \mathbf z_{t(i)}) \frac{q(\mathbf{\tilde{z}}_{s(i)})}{p(\mathbf{\tilde{z}}_{s(i)})}} \label{equ_wk10:solratio_zs}
\\&=
{\frac{p(\mathbf z_{s(i)} \mid \mathbf z_{t(i)}) \mathbb{E}_{\mathbf x_0 \sim p(\cdot \mid \mathbf{z}_{s(i)})}\Bigr[\frac{q(\mathbf x_0)}{p(\mathbf x_0)} \Bigl]}{\displaystyle\sum_{\mathbf{\tilde{z}}_{s(i)}}p(\mathbf{\tilde{z}}_{s(i)}\mid \mathbf z_{t(i)}) 
\mathbb{E}_{\mathbf x_0 \sim p(\cdot \mid \mathbf{\tilde{z}}_{s(i)})}\Bigr[\frac{q(\mathbf x_0)}{p(\mathbf x_0)} \Bigl]}} \label{equ_wk10:final}
\end{align}
For \ref{equ_wk10:sol_alpha} to \ref{equ_wk10:solratio_zs} we show that $\displaystyle\sum_{\mathbf x_0}\alpha(\mathbf x_0,\mathbf z_{t(i)})\,p(\mathbf z_{s(i)}\mid\mathbf z_{t(i)},\mathbf x_0) = p(\mathbf z_{s(i)} \mid \mathbf z_{t(i)}) \frac{q(\mathbf z_{s(i)})}{p(\mathbf z_{s(i)})}$: 
\begin{align}
    \displaystyle\sum_{\mathbf x_0}\alpha(\mathbf x_0,\mathbf z_{t(i)})\,p(\mathbf z_{s(i)}\mid\mathbf z_{t(i)},\mathbf x_0) &=  \displaystyle\sum_{\mathbf x_0}p(\mathbf x_{0}\mid\mathbf z_{t(i)}) \,\frac{q(\mathbf x_0)}{p(\mathbf x_0)}\,p(\mathbf z_{s(i)}\mid\mathbf z_{t(i)},\mathbf x_0)
    \\&=
    \displaystyle\sum_{\mathbf x_0} p(\mathbf z_{t(i)} \mid \mathbf x_{0}) \,\frac{p(\mathbf x_0)}{p(\mathbf z_{t(i)})}\,\frac{q(\mathbf x_0)}{p(\mathbf x_0)}\,\frac{p(\mathbf z_{t(i)} \mid \mathbf z_{s(i)})\cdot p (\mathbf z_{s(i)} \mid \mathbf x_{0})}{p (\mathbf z_{t(i)} \mid \mathbf x_{0})}
    \\&=
    \frac{p(\mathbf z_{t(i)} \mid\mathbf z_{s(i)})}{p(\mathbf z_{t(i)})}\displaystyle\sum_{\mathbf x_0}q(\mathbf x_0) p(\mathbf z_{s(i)} \mid \mathbf x_0)
    = p(\mathbf z_{t(i)} \mid\mathbf z_{s(i)})\frac{q(\mathbf z_{s(i)})}{p(\mathbf z_{t(i)})}
\end{align}
For \ref{equ_wk10:solratio_zs} to \ref{equ_wk10:final} we show that $\frac{q(\mathbf{z}_{s(i)})}{p(\mathbf{z}_{s(i)})}=\mathbb E_{\mathbf{x}_0\sim p(\cdot\mid \mathbf{z}_{s(i)})}\!\bigl[q(\mathbf{x}_0)/p(\mathbf{x}_0)\bigr]$:
\begin{align}
\frac{q(\mathbf{z}_{s(i)})}{p(\mathbf{z}_{s(i)})} &=  \frac{\sum_{\mathbf{x}_0}q(\mathbf{z}_{s(i)}\mid \mathbf{x}_0)\,q(\mathbf{x}_0)}
          {p(\mathbf{z}_{s(i)})} &
        \\&= \frac{\sum_{\mathbf{x}_0}q(\mathbf{z}_{s(i)}\mid \mathbf{x}_0)\,p(\mathbf{x}_0) \; r({\mathbf{x}_0})}
          {p(\mathbf{z}_{s(i)})}
        =\sum_{\mathbf{x}_0}\frac{p(\mathbf{z}_{s(i)}\mid \mathbf{x}_0)\,p(\mathbf{x}_0)}{p(\mathbf{z}_{s(i)})} \; r({\mathbf{x}_0})
        \\&=  \sum_{\mathbf{x}_0} p(\mathbf{x}_0 \mid \mathbf{z}_{s(i)})\; r({\mathbf{x}_0})
        =\mathbb E_{\mathbf{x}_0\sim p(\cdot\mid \mathbf{z}_{s(i)})}\!\bigl[q(\mathbf{x}_0)/p(\mathbf{x}_0)\bigr] \label{equ_apx:ratio}
\end{align}

\section{Guided $\tau$-Sampling}
\begin{algorithm}
\caption{\textsc{Naive Guided Denoise Step} $(\mathbf z_t,\ t,\ \Delta t)$}
\label{alg:naive_ratio_denoise}
\begin{algorithmic}[1]
\REQUIRE noisy sequence $\mathbf z_t \in \mathcal V^{L}$, time $t$, step $\Delta t>0$, denoiser $p_\theta$, ratio net $r_\phi$, mask index $m$, guidance schedule $\gamma(\cdot)$, noise schedule $\sigma(\cdot)$
\ENSURE updated sequence $\mathbf z_s$
\STATE $\sigma_t \gets \sigma(t)$
\STATE $\log x_\theta \gets p_\theta(\mathbf z_t,\ \sigma_t)$ \COMMENT{$L\times |\mathcal V|$ logits}
\STATE $E \gets \{\ell \in \{1,\ldots,L\} : z_t^{(\ell)} = m\}$ \COMMENT{masked positions}
\STATE $\mathbf z_s \gets \mathbf z_t$
\FORALL{$\ell \in E$} 
  \FORALL{$v \in \mathcal V$}
    \STATE $\tilde{\mathbf z} \gets \mathbf z_t$ 
    \STATE $\tilde{\mathbf z}^{(\ell)} \gets v$
    \STATE $\log r[\ell,v] \gets \log r_\phi(\tilde{\mathbf z},\ \sigma_t)$
  \ENDFOR
  \STATE $\log q^{\text{guided}}[\ell,:] \gets \log x_\theta[\ell,:] + \gamma(t)\,\log r[\ell,:]$
  \STATE $\pi^{(\ell)} \gets \operatorname{softmax}\!\big(\log q^{\text{guided}}[\ell,:]\big)$
  \STATE $z_s^{(\ell)} \sim \operatorname{Cat}\!\big(\pi^{(\ell)}\big)$
\ENDFOR
\STATE \RETURN $\mathbf z_s$
\end{algorithmic}
\end{algorithm}
In the experiments, we observe that the combination of top-$n_{\text{ratio}}$ selection and the guidance weight $\gamma$ can misallocate probability mass, potentially causing the diffusion process to collapse into a fully masked sequence at the final step. For instance, when the top-$n_{\text{ratio}}$ is small, the denoiser tends to propose tokens that are highly likely to belong to the source domain, which then receive negative logits from the guidance ratio network. This effect is especially pronounced in high-noise regions, where all tokens experience a reduction in their probability of being unmasked, though some are affected more than others. Consequently, the overall probability of unmasking decreases, while the probability of remaining masked is incorrectly increased. This leads to the issue of ending up with a higher number of masked positions in the final step. To mitigate this problem, we apply a position-independent normalization:
\begin{align*}
    q_{\theta,\phi}^\gamma(\mathbf z_s=\mathbf m \mid \mathbf z_t) = p(\mathbf z_s=\mathbf m \mid \mathbf z_t) = p_{\mathbf m}, \quad \quad
    \sum_{_{{\mathbf v} \in \mathcal V \setminus \{\mathbf m\}}} q_{\theta,\phi}^\gamma(\mathbf z_s=\mathbf v \mid \mathbf z_t) = 1 - p_{\mathbf m},
\end{align*}
which implies the renormalized update over non-mask tokens, thus stabilizing the denoising diffusion process.
\begin{align*}
    q_{\theta,\phi}^\gamma(\mathbf z_s=\mathbf v \mid \mathbf z_t)
    = \frac{(1-p_{\mathbf m})\, p_\theta(\mathbf z_s=\mathbf v \mid \mathbf z_t)\, r_\phi(\mathbf z_s=\mathbf v)^\gamma}
           {\displaystyle \sum_{\tilde{\mathbf v} \in \mathcal V \setminus \{\mathbf m\}}
            p_\theta(\mathbf z_s=\tilde{\mathbf v} \mid \mathbf z_t)\, r_\phi(\mathbf z_s=\tilde{\mathbf v})^\gamma}
    \quad \text{for all } \mathbf v \in \mathcal V \setminus \{\mathbf m\}.
\end{align*}
We use this algorithm in the first experiments section.
\section{Training Details} 
\textbf{The Code will be released upon acceptance.} In this section, we briefly describe the training setup for reproducibility. 
All training networks share the same learning schedule, which includes cosine decay with warmup and a high learning rate of $3 \times 10^{-4}$. The Adam optimizer is used, and we apply a dropout rate of 0.1. The batch size is set to 256 across all training algorithms.
The time-independent and time-dependent Classifier is trained with 0.1 label smoothing and 4000 gradient steps. Importantly, the classifier is trained with binary labels, where the target domain is labeled as zero and the source domain as one. The ratio network is trained with $\lambda$ set to 0.1, as outlined in the following pseudo-algorithm \ref{alg:ratio_TLDM_with_rglz}. 
\begin{algorithm}
\caption{Training loop for the ratio network $r_{\phi}(x_t,t)$}
\label{alg:ratio_TLDM_with_rglz}
\begin{algorithmic}[1]
\REQUIRE
  source data $p(x)$ and target data $q(x)$, frozen classifier $d_{\omega}: \mathcal V^L \rightarrow [0,1]$, \\ frozen time-dependent classifier $d_{\omega}:\mathcal V^L, t \rightarrow [0,1]$
  forward kernel \textsc{Corrupt}$(x_0,\sigma_t)$ with schedule $\sigma(t)$, \\
  batch size $b$, learning rate $\eta$, regularizer rate $\lambda$
\REPEAT
    \STATE \textbf{Sample} mini‑batch $x_0$ of size $b$ \textbf{from} $\mathbf p$ and \textbf{draw} $t\sim\text{Uniform}(0,1)$    
\STATE $x_t\gets\textsc{Corrupt}(x_0,\sigma(t))$                        \hfill  \COMMENT{forward diffusion}
    \STATE $\mathcal L_\text{guidance}\!=\!\dfrac1b\sum\|r_{\phi}(x_t,t) - \dfrac{1-d_{\omega}(x_0)}{d_{\omega}(x_0)}\|_2^2$             \hfill  \COMMENT{Guidance loss}
    \STATE
    \STATE \textbf{Sample} mini‑batch $x_0$ of size $b$ \textbf{from} $\mathbf q$ and \textbf{draw} $t\sim\text{Uniform}(0,1)$   
    \STATE $x_t\gets\textsc{Corrupt}(x_0,\sigma(t))$                       \hfill  \COMMENT{forward diffusion}
    \STATE $\mathcal L_\text{cycle}\!=\!\dfrac1b\sum\|r_{\phi}(x_t,t) - \dfrac{1-d_{\omega}(x_t,t)}{d_{\omega}(x_t,t)}\|_2^2$              \hfill \COMMENT{Guidance loss}
    \STATE 
    \STATE $\mathcal L = \mathcal L_\text{guidance} + \lambda \cdot  \mathcal L_\text{cycle}$
    \STATE $\psi\gets\psi-\eta\,\nabla_{\psi}\mathcal L$                   \hfill  \COMMENT{gradient update}
\UNTIL convergence
\STATE \RETURN trained parameters $\psi$
\end{algorithmic}
\end{algorithm}

For sample corruption, we use a log-linear schedule with $\sigma_{\text{max}} = 20$ and $\sigma_{\text{min}} = 0.0001$. The planner is trained following Algorithm~\ref{alg:planner_training}, using the same training configuration as described above, except that it is trained for 100k update steps. On the held-out validation set from the source domain, the planner achieves an average accuracy of 70\%. In high-noise regions, it becomes considerably more challenging for the planner to predict which positions will be correctly denoised, due to the high variability in token unmasking.

\begin{algorithm}
\caption{Training loop for the planner $\rho_{\vartheta}(x_t,t)$}
\label{alg:planner_training}
\begin{algorithmic}[1]
\REQUIRE
  training corpus $\mathcal D$, frozen source denoiser $x_\theta: \mathcal V^L \to \mathbb{R}^{L \times |\mathcal V|}$ (per-position logits/probs), \\
  forward kernel \textsc{Corrupt}$(x_0,\sigma(t))$ with schedule $\sigma(t)$, batch size $b$, learning rate $\eta$
\REPEAT
  \STATE \textbf{Sample} mini-batch $x_0$ of size $b$ from $\mathcal D$; draw $t \sim \mathrm{Uniform}(0,1)$
  \STATE $x_t \gets \textsc{Corrupt}(x_0,\sigma(t))$ \hfill \COMMENT{(1) corrupt the sample}
  \STATE $\log x_\theta \gets x_\theta(x_t)$;\;\; $\hat x_0 \gets \arg\max \log x_\theta$ \hfill \COMMENT{(2) pass to source denoiser}
  \STATE $M \gets (x_t = \texttt{[MASK]})$;\;\; $y \gets \mathbf{1}\!\left[\hat x_0 = x_0\right]$ \hfill \COMMENT{(3) where denoiser is correct (masked positions)}
  \STATE $s \gets \rho_{\vartheta}(x_t)$ \hfill \COMMENT{(4) planner logits per position}
  \STATE $\mathcal L \gets \mathrm{BCEWithLogits}\!\left(s[M],\, y[M]\right)$ \hfill \COMMENT{(5) loss over masked positions only}
  \STATE $\vartheta \gets \vartheta - \eta \nabla_{\vartheta}\mathcal L$ \hfill \COMMENT{gradient update}
\UNTIL convergence
\STATE \RETURN trained planner parameters $\vartheta$
\end{algorithmic}
\end{algorithm}
\subsection{Implementation on Markov Chain}
\label{sec:ImplementationMC}
We use the sampling process of \citep{sahoo2024simpleeffectivemaskeddiffusion} for the source, finetuned, and target discrete diffusion models. The backbone is a Diffusion Transformer (DiT) \citep{peebles2023scalablediffusionmodelstransformers} with about five million learnable parameters. The ratio estimator and the classifier use about half as many parameters with a final mean and fully connected layer that maps the output to a scalar. The exact implementation appears in our codebase. For guided sampling, we follow the ancestral sampling method. Since the vocabulary is small we set $n_{\text{ratio}}=N$ and use $20$ sampling steps. We train a masked diffusion model on the source domain for $30$ epochs. We then compare a vanilla masked diffusion model trained directly on the target samples (60 epochs), a finetuned diffusion model (90 epochs), and our GTL method (60 epochs). The finetuned model and our method share the same source diffusion model.
\section{Application to continuous-time Score-based Models}

\subsection{Background}
\citet{campbell2022continuoustimeframeworkdiscrete} introduces a continuous-time framework for discrete denoising models formulated as a continuous-time Markov chain (CTMC). Unlike the discrete-time setting of \citep{austin2023structureddenoisingdiffusionmodels} with steps $t=0,1,\dots,T$, here $t\in[0,T]$ is continuous.

In this section, we switch notation. The forward kernel is
$p_{t|s}(\mathbf y' \mid \mathbf y) := p(\mathbf z_t=\mathbf y' \mid \mathbf z_s=\mathbf y)$,
the backward kernel is
$p_{s|t}(\mathbf y' \mid \mathbf y) := p(\mathbf z_s=\mathbf y' \mid \mathbf z_t=\mathbf y)$,
the noising kernel is
$p_t(\mathbf y \mid \mathbf x) := p(\mathbf z_t=\mathbf y \mid \mathbf x_0=\mathbf x)$,
and the true backward transition is
$p(\mathbf y' \mid \mathbf y,\mathbf x) := p(\mathbf z_s=\mathbf y' \mid \mathbf z_t=\mathbf y,\mathbf x_0=\mathbf x)$.
This CTMC formulation allows transitions at arbitrary times. The forward corruption is specified by a rate matrix
$R_t \in \mathbb{R}^{|\mathcal V|\times|\mathcal V|}$, and the reverse process by a reverse rate matrix $\widetilde{R}_t$:
\begin{align*}
p_{t|s}(\mathbf y' \mid \mathbf y) &= \delta_{\mathbf y',\mathbf y} + R_t(\mathbf y',\mathbf y)\,\Delta t + \mathcal{O}\!\left(\Delta t^2\right),\\
p_{s|t}(\mathbf y' \mid \mathbf y,\mathbf x) &= \delta_{\mathbf y',\mathbf y} + \widetilde{R}_t(\mathbf y',\mathbf y)\,\Delta t + \mathcal{O}\!\left(\Delta t^2\right).
\end{align*}
Here $\Delta t$ denotes a small time increment when discretizing the continuous-time process, with $s=t-\Delta t$. Here, $\delta_{i,j}$ denotes the Kronecker delta (equal to $1$ if $i=j$ and $0$ otherwise), and $\mathcal{O}(\Delta t^{2})$ denotes terms of order $\Delta t^{2}$ or higher. The reverse rate can be written in terms of the forward rate and the marginals $q_t(\cdot\mid\mathbf x)$:
\[
\widetilde{R}_t(\mathbf y',\mathbf y)=
\begin{cases}
\dfrac{q_t(\mathbf y' \mid \mathbf x)}{q_t(\mathbf y \mid \mathbf x)}\, R_t(\mathbf y,\mathbf y'), & \mathbf y'\neq \mathbf y,\\[6pt]
-\displaystyle\sum_{\tilde{\mathbf y}\neq \mathbf y}
\dfrac{q_t(\tilde{\mathbf y} \mid \mathbf x)}{q_t(\mathbf y \mid \mathbf x)}\, R_t(\mathbf y,\tilde{\mathbf y}), & \mathbf y'=\mathbf y.
\end{cases}\]
\paragraph{Concrete score.}
We can express the reverse process in terms of the forward transition rates and a \emph{concrete score} defined by
$s_\theta(\mathbf y)_{\mathbf y'} \approx \tfrac{q_t(\mathbf y' \mid \mathbf x)}{q_t(\mathbf y \mid \mathbf x)}$, parameterized by a neural network $s_\theta : \mathcal V \to \mathbb R_{>0}^{|\mathcal V|}$. This enables training via concrete score matching (CSM)~\citep{meng2023concretescorematchinggeneralized}, directly analogous to score-based models in continuous domains~\citep{song2021scorebasedgenerativemodelingstochastic}. \citet{lou2024discretediffusionmodelingestimating} propose the diffusion-weighted denoising score entropy (DWDSE) loss (their CDM-based objective) and note it can also be derived from the continuous-time likelihood framework of \citep{campbell2022continuoustimeframeworkdiscrete}; a detailed derivation appears in Appendix C.3 of \citep{sahoo2024simpleeffectivemaskeddiffusion}.
Concretely,
\begin{align}
&\lim_{T \to \infty}
\mathbb{E}_{\substack{t \in \{\tfrac1T,\tfrac2T,\ldots,1\}\\ \mathbf y \sim q_t(\cdot \mid \mathbf x)}}
\!\bigl[\mathrm{D}_{\mathrm{KL}}\!\bigl(p_{s\mid t}(\mathbf y' \mid \mathbf y,\mathbf x)
\;\|\;
p_{s\mid t}(\mathbf y' \mid \mathbf y)\bigr)\bigr]
\label{equ_apdx:d3pm_loss}
\\&=
\mathbb{E}_{\substack{t \in [0,1]\\ \mathbf y \sim q_t(\cdot \mid \mathbf x)}}
\!\Bigl[
\sum_{\mathbf y' \neq \mathbf y}
R_t(\mathbf y,\mathbf y')
\Bigl(
s_\theta(\mathbf y)_{\mathbf y'}
-
\frac{q_t(\mathbf y' \mid \mathbf x)}{q_t(\mathbf y \mid \mathbf x)}
\log s_\theta(\mathbf y)_{\mathbf y'}
+
K\!\Bigl(\tfrac{q_t(\mathbf y' \mid \mathbf x)}{q_t(\mathbf y \mid \mathbf x)}\Bigr)
\Bigr)
\Bigr],
\label{equ_apdx:sedd_loss}
\end{align}
where $K(a) = a \log a -a $. 
\subsection{Theorem}
In the previous section, we introduced the diffusion process and the source-domain objective on which the score function $s_\theta$ is trained. We now state a theorem that enables sampling from the target distribution \emph{without} retraining $s_\theta$ under the objective in \ref{equ_apdx:sedd_loss}. Let $r_\phi(\mathbf y)$ denote the learned density-ratio guidance term defined earlier.

\begin{theorem}[Score-Based Ratio Transfer Learning] \label{Theorem:score}
Let $s_\theta(\mathbf y)_{\mathbf y'}$ be a score-based model trained on the source distribution $p$, and let $\widetilde R_t^\theta(\cdot,\cdot)$ be the reverse-rate matrix induced by $s_\theta$. Then, sampling from the target distribution $q$ is given by the target reverse-rate matrix:
\begin{equation}
    \widetilde R_t^{\psi}(\mathbf y',\mathbf y)=
\begin{cases}
\dfrac{r_\phi(\mathbf y')}{r_\phi(\mathbf y)}\,\widetilde R_t^\theta(\mathbf y',\mathbf y), & \mathbf y'\neq \mathbf y,\\[8pt]
-\displaystyle\sum_{\tilde{\mathbf y}\neq \mathbf y}\dfrac{r_\phi(\tilde{\mathbf y})}{r_\phi(\mathbf y)}\,\widetilde R_t^\theta(\tilde{\mathbf y},\mathbf y), & \mathbf y'=\mathbf y.
\end{cases}
\end{equation}
\end{theorem}
\subsection{Proof of Theorem \ref{Theorem:score}}
For the proof, we take the loss for a small step $\Delta t$ with $s=t-\Delta t$ from \eqref{equ_apdx:d3pm_loss} and apply Theorem~\ref{theorem1}, which yields the following expression:
\begin{align}
q_{s\mid t}^{\psi^\star}(\mathbf{y}'\mid\mathbf{y})
&=
\frac{
p_{s\mid t}(\mathbf{y}'\mid\mathbf{y})\,
\mathbb{E}_{\mathbf{x}\sim p(\cdot\mid\mathbf{y}')}\!
\bigl[\tfrac{q(\mathbf{x})}{p(\mathbf{x})}\bigr]
}{
\displaystyle
\sum_{\widetilde{\mathbf{y}}}
p_{s\mid t}(\widetilde{\mathbf{y}}\mid\mathbf{y})\,
\mathbb{E}_{\mathbf{x}\sim p(\cdot\mid\widetilde{\mathbf{y}})}\!
\bigl[\tfrac{q(\mathbf{x})}{p(\mathbf{x})}\bigr]
}
\\&=
\frac{p_{s\mid t}(\mathbf y'\mid\mathbf y)\,r_\phi(\mathbf y')}
     {\displaystyle\sum_{\widetilde{\mathbf y}}
       p_{s\mid t}(\widetilde{\mathbf y}\mid\mathbf y)\,r_\phi(\widetilde{\mathbf y})}
\end{align}
We can already assume that the source backward kernel is approximated by $p_{s\mid t}(\mathbf y' \mid \mathbf y)
= \delta_{\mathbf y',\mathbf y} + \widetilde R_t^\theta(\mathbf y',\mathbf y)\,\Delta t + \mathcal O(\Delta t^2)$: 
\[
q_{s\mid t}^{\psi}(\mathbf y' \mid \mathbf y)
= \frac{\bigl[\delta_{\mathbf y',\mathbf y} + \widetilde R_t^\theta(\mathbf y',\mathbf y)\Delta t\bigr]\,r_\phi(\mathbf y')}
       {\sum_{\tilde{\mathbf y}}\bigl[\delta_{\tilde{\mathbf y},\mathbf y} + \widetilde R_t^\theta(\tilde{\mathbf y},\mathbf y)\Delta t\bigr]\,r_\phi(\tilde{\mathbf y})} \;+\;\mathcal O(\Delta t^2)
\]
We simplify the denominator by using the Taylor expansion of $1/(1+v)$ around $v=0$:
$[\frac{1}{1+v} = 1 - v + \mathcal{O}(v^2)]$ and replacing $S:=\sum_{\tilde{\mathbf y}}\widetilde R_t^\theta(\tilde{\mathbf y},\mathbf y)\,r_\phi(\tilde{\mathbf y})$:
\begin{align*}
    \frac{1}{\sum_{\tilde{\mathbf y}}\bigl[\delta_{\tilde{\mathbf y},\mathbf y} + \widetilde R_t^\theta(\tilde{\mathbf y},\mathbf y)\Delta t\bigr]\,r_\phi(\tilde{\mathbf y})} = \frac{1}{r_\phi(\mathbf y)+\Delta t\,S}
= \frac{1}{r_\phi(\mathbf y)}\Bigl(1+\frac{\Delta t\,S}{r_\phi(\mathbf y)}\Bigr)^{-1}
= \frac{1}{r_\phi(\mathbf y)}-\frac{\Delta t\,S}{r_\phi(\mathbf y)^2}+\mathcal O(\Delta t^2)
\end{align*}
\begin{enumerate}
    \item For the Off–diagonal case $\mathbf y'\neq \mathbf y$, hence $\delta_{\mathbf y',\mathbf y}=0$: 
    \begin{align*}
        q_{s\mid t}^{\psi}(\mathbf y' \mid \mathbf y)
        &= \frac{\bigl[\delta_{\mathbf y',\mathbf y} + \widetilde R_t^\theta(\mathbf y',\mathbf y)\Delta t\bigr]\,r_\phi(\mathbf y')}
       {\sum_{\tilde{\mathbf y}}\bigl[\delta_{\tilde{\mathbf y},\mathbf y} + \widetilde R_t^\theta(\tilde{\mathbf y},\mathbf y)\Delta t\bigr]\,r_\phi(\tilde{\mathbf y})} \;+\;\mathcal O(\Delta t^2)
        \\&= \widetilde R_t^\theta(\mathbf y',\mathbf y)\,r_\phi(\mathbf y')\,\Delta t
        \left(\frac{1}{r_\phi(\mathbf y)}-\frac{\Delta t\,S}{r_\phi(\mathbf y)^2}+\mathcal O(\Delta t^2)\right)
        \\&= \frac{\widetilde R_t^\theta(\mathbf y',\mathbf y)\,r_\phi(\mathbf y')}{r_\phi(\mathbf y)}\,\Delta t
        \;+\;\mathcal O(\Delta t^2),
    \end{align*}
    \item For the diagonal case $\mathbf y'=\mathbf y$:
    \begin{align*}
        q_{s\mid t}^{\psi}(\mathbf y \mid \mathbf y)
&= \frac{\bigl[1+\widetilde R_t^\theta(\mathbf y,\mathbf y)\,\Delta t\bigr]\,r_\phi(\mathbf y)}
       {r_\phi(\mathbf y)+\Delta t\,S}
+ \mathcal O(\Delta t^2)
= \frac{1+\widetilde R_t^\theta(\mathbf y,\mathbf y)\,\Delta t}{1+\frac{\Delta t\,S}{r_\phi(\mathbf y)}}+\mathcal O(\Delta t^2)
\\&= \Bigl(1+\widetilde R_t^\theta(\mathbf y,\mathbf y)\,\Delta t\Bigr)\Bigl(1-\frac{\Delta t\,S}{r_\phi(\mathbf y)}\Bigr)+\mathcal O(\Delta t^2)
= 1 + \Bigl[\widetilde R_t^\theta(\mathbf y,\mathbf y)-\frac{S}{r_\phi(\mathbf y)}\Bigr]\Delta t + \mathcal O(\Delta t^2).
    \end{align*}
\end{enumerate}
Putting these two cases together, we arrive at the following:
\begin{align*}
    q_{s\mid t}^{\psi}(\mathbf y' \mid \mathbf y)&= \delta_{\mathbf y',\mathbf y}\Bigl[1+\bigl(\widetilde R_t^\theta(\mathbf y,\mathbf y)-\sum_{\tilde{\mathbf y}}\widetilde R_t^\theta(\tilde{\mathbf y},\mathbf y)\,\tfrac{r_\phi(\tilde{\mathbf y})}{r_\phi(\mathbf y)}\bigr)\Delta t\Bigr]  +(1-\delta_{\mathbf y',\mathbf y})\Bigl[\widetilde R_t^\theta(\mathbf y',\mathbf y)\,\tfrac{r_\phi(\mathbf y')}{r_\phi(\mathbf y)}\,\Delta t\Bigr] + \mathcal O(\Delta t^2)
    \\&= \delta_{\mathbf y',\mathbf y}+{\Delta t}\Bigl[\widetilde R_t^\theta(\mathbf y',\mathbf y)\,\frac{r_\phi(\mathbf y')}{r_\phi(\mathbf y)} - \delta_{\mathbf y',\mathbf y}\,\sum_{\tilde{\mathbf y}} {\widetilde R_t^\theta(\tilde{\mathbf y},\mathbf y)\,\frac{r_\phi(\tilde{\mathbf y})}{r_\phi(\mathbf y)}} \Bigr] + \mathcal O(\Delta t^2),
    \\&=\delta_{\mathbf{y}',\mathbf y} + \Delta t\,\widetilde R_t^\psi(\mathbf y’,\mathbf y)\,
+\mathcal O(\Delta t^2).
\end{align*}
Thus concluding the proof.
\[
\widetilde R_t^\psi(\mathbf y',\mathbf y)
:= \widetilde R_t^\theta(\mathbf y',\mathbf y)\,\frac{r_\phi(\mathbf y')}{r_\phi(\mathbf y)}
- \delta_{\mathbf y',\mathbf y}\,\frac{1}{r_\phi(\mathbf y)}
\sum_{\tilde{\mathbf y}}\widetilde R_t^\theta(\tilde{\mathbf y},\mathbf y)\,r_\phi(\tilde{\mathbf y}).
\]
The work from \citep{nisonoff2025unlockingguidancediscretestatespace} in equation (2) arrives at a similar term just for classifier-based guidance.

\end{document}